\documentclass[10pt,twocolumn,letterpaper]{article}

\usepackage{wacv}
\usepackage{times}
\usepackage{epsfig}
\usepackage{graphicx}
\usepackage{amsmath}
\usepackage{amssymb}
\usepackage{booktabs}
\usepackage{pifont}
\usepackage{multirow}
\usepackage{adjustbox}
\usepackage{caption}
\usepackage{inconsolata}
\usepackage{comment}
\usepackage{breqn}
\usepackage{balance}

\usepackage[accsupp]{axessibility}  %

\usepackage{lipsum}

\usepackage{breqn}

\newcommand\blfootnote[1]{%
  \begingroup
  \renewcommand\thefootnote{}\footnote{#1}%
  \addtocounter{footnote}{-1}%
  \endgroup
}

\def\wacvPaperID{832} %

\wacvfinalcopy %

\ifwacvfinal
\usepackage[pagebackref=true,colorlinks,breaklinks=true,bookmarks=false]{hyperref}
\else
\usepackage[pagebackref=true,breaklinks=true,colorlinks,bookmarks=false]{hyperref}
\fi

\newcommand{\cmark}{\ding{51}}%
\newcommand{\xmark}{\ding{55}}%

\renewcommand{\paragraph}[1]{\vspace{1mm}\noindent\textbf{#1}}

\begin{document}

\title{Unsupervised Audio-Visual Lecture Segmentation}

\author{
Darshan Singh S$^{*}$ \hspace{0.3cm}
Anchit Gupta$^{*}$ \hspace{0.3cm}
C. V. Jawahar \hspace{0.3cm}
Makarand Tapaswi \\
CVIT, IIIT Hyderabad \\
\url{https://cvit.iiit.ac.in/research/projects/cvit-projects/avlectures}
}

\maketitle

\begin{abstract}
Over the last decade, online lecture videos have become increasingly popular and have experienced a meteoric rise during the pandemic.
However, video-language research has primarily focused on instructional videos
or movies, and tools to help students navigate the growing online lectures are lacking.
Our first contribution is to facilitate research in the educational domain by introducing \emph{AVLectures}, a large-scale dataset consisting of 86 courses with over 2,350 lectures covering various STEM subjects.
Each course contains video lectures, transcripts, OCR outputs for lecture frames, and optionally lecture notes, slides, assignments, and related educational content that can inspire a variety of tasks.
Our second contribution is introducing \emph{video lecture segmentation} that splits lectures into bite-sized topics.
Lecture clip representations leverage visual, textual, and OCR cues and are trained on a pretext self-supervised task of matching the narration with the temporally aligned visual content.
We formulate lecture segmentation as an unsupervised task and use these representations to generate segments using a temporally consistent 1-nearest neighbor algorithm, TW-FINCH~\cite{sarfraz2021temporally}.
We evaluate our method on 15 courses and compare it against various visual and textual baselines, outperforming all of them. 
Our comprehensive ablation studies also identify the key factors driving the success of our approach.

\end{abstract}

\vspace{-2mm}
\section{Introduction}
\vspace{-2mm}
\blfootnote{* indicates equal first author contribution}
\begin{figure*}
\centering
\includegraphics[width=0.90\linewidth]{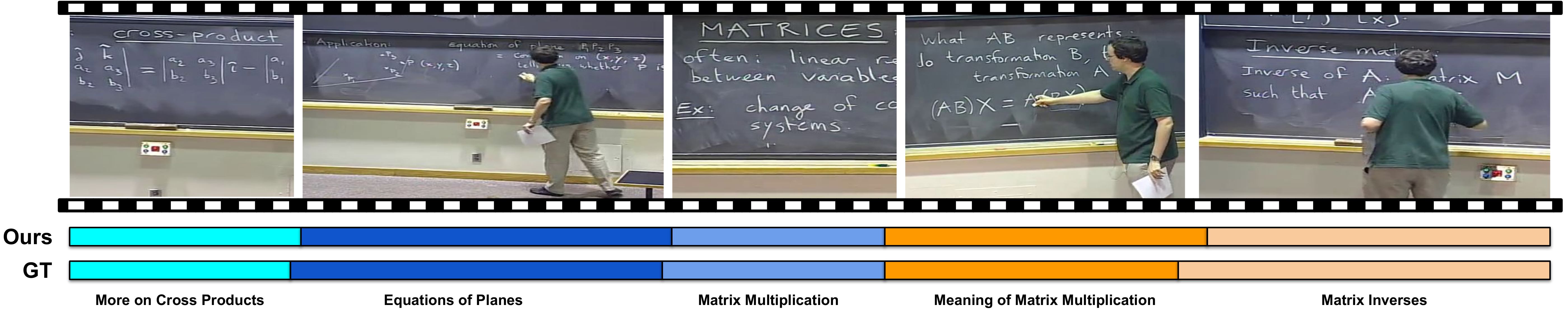}
\vspace{-2mm}
\caption{We address the task of lecture segmentation in an unsupervised manner. We show an example of a lecture segmented using our method. Our method predicts segments close to the ground-truth. Note that our method \emph{does not predict the segment labels}, they are only shown so that the reader can appreciate the different topics.}
\label{fig:banner}
\vspace{-4mm}
\end{figure*}

The last decade has seen a significant increase in online lectures in the form of Massive Open Online Courses (MOOCs) through platforms such as Coursera or EdX.
Many high-quality recorded lectures are also published online, \eg, MIT through MIT OpenCourseWare (OCW)\footnote{MIT-OCW - \url{https://ocw.mit.edu/}}, top Indian universities through NPTEL\footnote{NPTEL - \url{https://nptel.ac.in/}}, and several professors that make their lectures publicly available\footnote{\eg ~\href{https://youtube.com/playlist?list=PL2SOU6wwxB0uwwH80KTQ6ht66KWxbzTIo}{Statistics 110} or \href{https://www.youtube.com/playlist?list=PLC1qU-LWwrF64f4QKQT-Vg5Wr4qEE1Zxk}{Stanford's CS231n}.}.
This increase in online content is considered one of the biggest turning points in the history of education as anybody can learn any topic from the world's leading teachers from the comfort of their home~\cite{oer, gandhi2015topic}.
As the world moved to an online mode during the pandemic, there is absolutely no doubt that such online lecture content creation will only increase.

Creating an online course requires tremendous effort from the instructor and teaching assistants.
Apart from designing and preparing the content itself, the mode of presentation poses challenges including segmenting the large videos into smaller topics to enhance the learning experience, adding quiz-like questions during the lecture to retain the student’s engagement, summarizing the lecture at the end, \etc.
These tasks require carefully combing through the lecture several times, a time-consuming and error-prone process.
Our goal is to encourage the community to address these tasks automatically or at least provide automatic recommendations for a human-in-the-loop system as they have the potential to reduce instructor's efforts, giving them more time and energy to improve the lecture content.

To build such solutions, machine understanding of audio-visual (AV) lectures is crucial.
However, currently, there are no large-scale datasets of audio-visual lectures\footnote{Despite educational videos being the fourth most consumed content on the Internet according to \href{https://www.oberlo.com/statistics/online-video-consumption-statistics}{this survey}, just behind ``How-to" videos.}. 
Our \emph{first contribution} is \emph{AVLectures}, a large-scale dataset to facilitate research in automatic understanding of lecture videos (see Sec.~\ref{sec:avld} for details and statistics).
By releasing \emph{AVLectures}, we wish to ignite research in the largely overlooked applications in education to help manage the fast-growing online lecture content.

Our \emph{second contribution} is the formulation and benchmarking of the \emph{lecture segmentation} task, where, given a long video lecture, our goal is to temporally segment it into smaller bite-sized topics.
Lecture segmentation can be more challenging than scene segmentation in movies~\cite{rao2020movienetscenes} or cooking videos~\cite{kukleva2019unsupervised} as the differences across segments are subtle, in both the visual and transcribed narrations.
For example, Fig.~\ref{fig:banner} shows a professor teaching on the blackboard and walking along the podium.
A model trained on movies or instructional videos may find it hard to segment the lecture as the objects or actions in the video do not change significantly.
Across segments, the visual boundaries are subtle changes such as clearing the board, while the narration may see a shift in the overall topic of discussion.

We propose lecture segmentation as an unsupervised task that leverages visual, textual, and OCR cues from the audio-visual lecture.
We first split the lecture into small clips and extract each clip's visual and textual features using pre-trained models. 
To make our representations lecture-aware, we learn a joint text-video embedding in a self-supervised manner by matching the narration with the aligned visual content. 
Finally, we obtain clusters using a temporally consistent\footnote{Temporally \emph{consistent} here refers to temporally \emph{contiguous}, \ie~the segment membership of clips looks like [0, 0, 1, 1, 1, 2, 2] rather than [0, 1, 0, 2, 2, 1, 1].
TW-FINCH~\cite{sarfraz2021temporally} allows this over base FINCH~\cite{sarfraz2019efficient}.} 1-nearest neighbor algorithm,
TW-FINCH~\cite{sarfraz2021temporally}.

We pick lecture segmentation as our first use case based on an insightful large-scale study conducted on the EdX platform~\cite{guo2014demographic}.
They find that students who successfully complete an online course typically spend 4.4 minutes on a 12-15 minute long lecture clip, clearly demonstrating the need for simplified navigation of long clips.
Lecture segmentation is also a first step towards creating a multimodal table of contents to summarize a lecture~\cite{mahapatra2018videoken}.
Finally, there is evidence for segmentation to assist in enabling non-linear video consumption~\cite{verma2021non} and efficient previewing~\cite{bulathwela2020s, chen2018temporally, perez2021x5learn}.
While segmentation is our first task, we emphasize that \emph{AVLectures} can be used for various other tasks in the future such as generating automatic quizzes for the lecture, aligning lecture videos with the notes enabling generation of lecture notes, retrieving relevant clips of the lecture using text queries, summarizing long lecture videos, retrieving and aligning similar courses/lectures from different learning platforms, and many more.

Our key contributions are summarized below.
(i) We introduce a novel educational audio-visual lectures dataset, \emph{AVLectures}, that can facilitate several applications in the education domain.
(ii) We formulate and benchmark the problem of \emph{unsupervised lecture segmentation}.
We show that self-supervised multimodal representations learned by matching the narration with temporally aligned video clips greatly help the task of segmentation.
(iii) Our method outperforms several baselines. We also provide extensive ablation studies to understand prominent factors leading to the success of our approach.
We will release code and data.

\section{Related Work}

\begin{figure*}[t]
\centering
\includegraphics[width=0.98\linewidth]{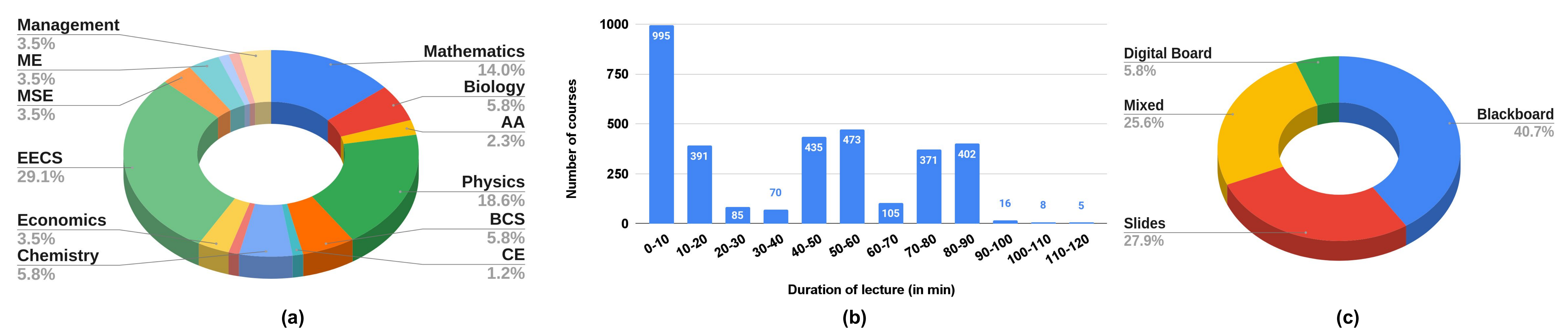}
\vspace{-2mm}
\caption{AVLectures statistics.
(a) \textbf{Subject areas.}
ME: Mechanical Eng.,
MSE: Materials Science and Eng.,
EECS: Electrical Eng. and Computer Science,
AA: Aeronautics and Astronautics,
BCS: Brain and Cognitive Sciences,
CE: Chemical Eng.
(b) \textbf{Lecture duration} distribution.
(c) \textbf{Presentation modes} distribution.}
\vspace{-3mm}
\label{fig:dataset}
\end{figure*}

\paragraph{Applications in educational videos.}
Research in video-language domain has focused primarily on movies~\cite{papalampidi2020movie, rohrbach2017movie, tapaswi2016movieqa}, and instructional videos~\cite{alayrac2016unsupervised, miech2019howto100m, rouditchenko2020avlnet}, especially cooking videos~\cite{youcook, zhou2018weakly}.
However, there are a few isolated works~\cite{bulathwela2021peek, bulathwela2020vlengagement, dutta2018localizing, gandhi2015topic, mahapatra2018automatic, mahapatra2018videoken} that attempt to solve various problems in the education domain that we highlight below.
Mahapatra~\etal~\cite{mahapatra2018automatic} propose an approach to generate a hierarchical table of contents for a lecture video using multimodal information such as transcripts and associated metadata from video key frames.
In the direction of localizing and recognizing text on a blackboard, Dutta~\etal~\cite{dutta2018localizing} introduce LectureVideoDB, a dataset consisting of frames from multiple lecture videos (including blackboard).
Bulathwela~\etal~\cite{bulathwela2021peek,bulathwela2020vlengagement} introduce datasets to understand learner engagement with educational videos.

Related to our work, lecture video segmentation was first proposed by Gandhi~\etal~\cite{gandhi2015topic}.
A visual saliency algorithm is adopted to find the topic transition points in the lecture automatically, however, this works primarily for slide-based lectures.
In contrast, our method shows promising results across all lecture types: blackboard, slide-based, and digital board. 
Additionally, the dataset of~\cite{gandhi2015topic} is orders of magnitude smaller, 10 vs. 2,350 lectures.
Finally, AVLectures is not only video material but is augmented by rich metadata, including transcripts, OCR outputs for slides/blackboard frames, lecture notes, lecture slides, and assignments.

\paragraph{Joint representation learning of video and language.}
Our proposed model learns meaningful representations of lectures and aligned transcripts, which we use to perform the lecture segmentation task.
In this section, we review popular works that address joint representation learning in video and language.
A common self-supervised objective used to learn good representations is aligning video with its corresponding narrations~\cite{miech2020milnce,miech2019howto100m}, which can then be used for a number of downstream tasks, such as
text-to-video retrieval~\cite{fu2021violet, lei2021less, miech2019howto100m},
visual question answering~\cite{VQA,tapaswi2016movieqa,yang2021justask},
video captioning~\cite{iashin2020multi, pan2016hierarchical, yu2016video}, 
natural language guided video summarization~\cite{narasimhan2021clip} among others.
Typically, representations from off-the-shelf pre-trained visual and language models are improved via a joint video-text embedding trained on the alignment task~\cite{miech2019howto100m}.
Recent approaches~\cite{desai2021virtex,fu2021violet, lei2021less} also adopt Transformer-based models that learn in an end-to-end manner from raw video pixels.
Our work explores the first direction.
We extract video features using off-the-shelf models and combine them with OCR features.
Then joint embeddings are learned using a pretext self-supervised task of matching the embeddings from narrations with temporally aligned video clips. 

\paragraph{Temporal video segmentation.} 
While fully supervised~\cite{ding2017tricornet}, 
weakly supervised~\cite{li2019weakly, souri2019weakly},
and unsupervised~\cite{aakur2019perceptual, alayrac2016unsupervised, kukleva2019unsupervised, sarfraz2021temporally} approaches have been explored,
we adopt the unsupervised path as collecting ground-truth segmentation labels is challenging, and we would like our method to generalize to diverse courses from novel educational platforms.
In the unsupervised space, instructional videos are segmented by finding and grouping direct object relations in the narrations~\cite{alayrac2016unsupervised}
or through the use of frame-level features that incorporates relative temporal information followed by K-means clustering (CTE)~\cite{kukleva2019unsupervised}.
Proxy tasks such as future frame prediction are also used to perform temporal segmentation~\cite{aakur2019perceptual}.
Recently, a temporally weighted version of a 1-nearest neighbor clustering algorithm is proposed to produce temporally consistent clusters (TW-FINCH)~\cite{sarfraz2021temporally}.
We will show that self-supervised joint text-video representation learning together with TW-FINCH leads to good segmentation performance on AVLectures.

\section{The AVLectures Dataset}
\label{sec:avld}

We introduce \emph{AVLectures}, a large-scale educational audio-visual lectures dataset to facilitate research in the domain of lecture video understanding.
The dataset comprises of 86 courses with over 2,350 lectures for a total duration of 2,200 hours.
Each course in our dataset consists of video lectures, corresponding transcripts, OCR outputs for frames, and optionally lecture notes, slides, and other metadata, making our dataset a rich multi-modality resource.

Courses span a broad range of subjects, including Mathematics, Physics, EECS, and Economics (see Fig.~\ref{fig:dataset}a).
While the average duration of a lecture in the dataset is about 55 minutes, Fig.~\ref{fig:dataset}b shows a significant variation in the duration.
We broadly categorize lectures based on their presentation modes into four types:
(i)~Blackboard, (ii)~Slides, (iii)~Digital Board, and (iv)~Mixed, a combination of blackboard and slides.
Fig.~\ref{fig:dataset}c depicts a healthy distribution of presentation modes in our dataset.
Additional statistics are presented in the supplementary material.

\paragraph{Courses with Segmentation.}
Among the 86 courses in AVLectures, a significant subset of 15 courses also have temporal segmentation boundaries.
We refer to this subset as the \emph{Courses with Segmentation} (CwS) and the remainder 71 courses as the \emph{Courses without Segmentation} (CwoS).

\begin{figure*}[t]
\includegraphics[width=\linewidth]{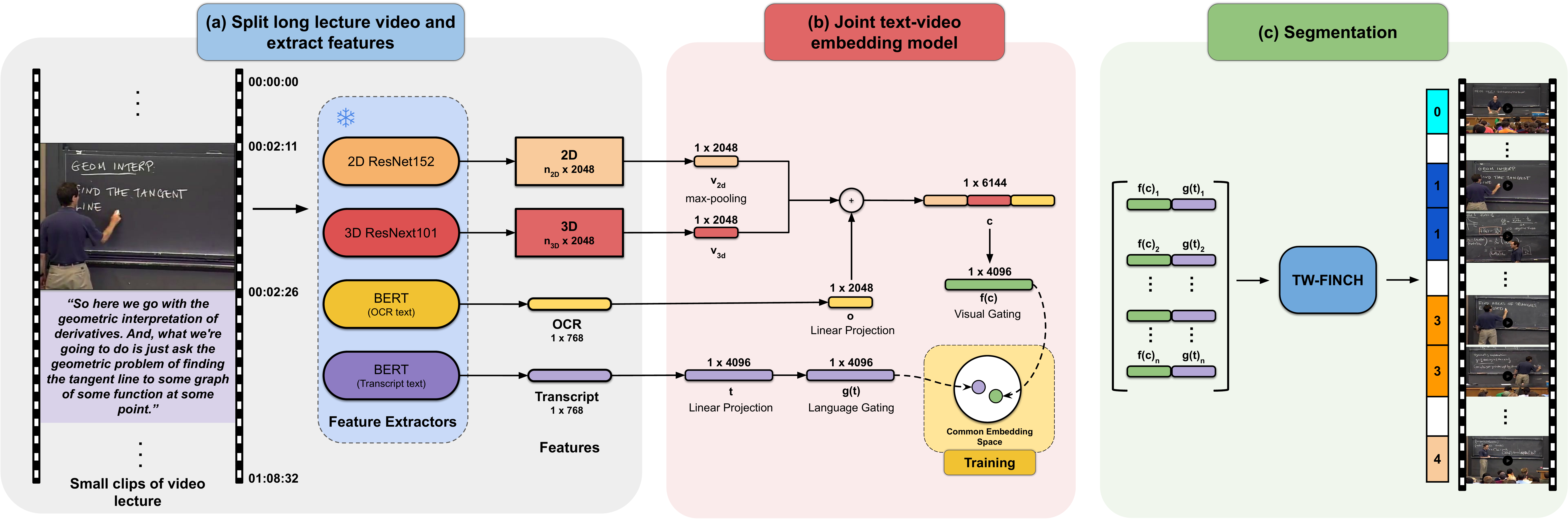}
\vspace{-5mm}
\caption{\textbf{Segmentation pipeline.}
(a)~\emph{Video clip and feature extraction pipeline} used to extract visual and textual features from small clips of 10s-15s duration.
The feature extractors are frozen and are not fine-tuned during the training process.
(b)~\emph{Joint text-video embedding model} learns lecture-aware representations.
(c)~\emph{Lecture segmentation process}, where we apply TW-FINCH at a clip-level to the learned (concatenated) visual and textual embeddings obtained from (b).}
\vspace{-3mm}
\label{fig:banner_2}
\end{figure*}

\subsection{Dataset Collection Procedure}
\label{subsec:datacollection}
Our dataset is primarily sourced from MIT-OCW~\cite{mit_ocw}.
We curated a list of courses by browsing the OCW website and used web scraping tools to download the video lectures and accompanying metadata such as narration transcripts, assignments, lecture notes/slides, \etc.
Non-lecture videos (\eg~instructor interviews) that were found in some courses are manually discarded.
We process and store the OCR outputs of video frames in each lecture using Google Cloud Vision API.
As sudden changes in the visual content of a lecture are rare, we process one frame at every 10 seconds.

\subsection{Curating the Lecture Segmentation Dataset}
It is shown that partitioning a long duration lecture into shorter topic-based clips helps in capturing students' attention and improves the overall learning experience~\cite{guo2014demographic, verma2021non}.
However, manually segmenting lecture recordings is a time-consuming and costly task.
To evaluate automatic methods for lecture segmentation,
we create a subset of our dataset, called \emph{Courses with Segmentation} (CwS), that includes courses in which long lecture videos are segmented into multiple smaller clips.
We curate 15 such courses with 350 lectures in total, where temporal segmentation \emph{ground-truth} (for each lecture) is obtained in one of two ways.
(i)~Out of the 15 courses, 5 courses\footnote{(i) \eg~\href{https://ocw.mit.edu/courses/18-01sc-single-variable-calculus-fall-2010/pages/1.-differentiation/part-a-definition-and-basic-rules/session-1-introduction-to-derivatives/}{Single Variable Calculus}} have topics in the table of contents that refer to various temporal segments in a long lecture video.
We obtain the segmentation timestamps for such courses directly by web scraping.
(ii)~The rest of the 10 courses\footnote{(ii) \eg~\href{https://ocw.mit.edu/courses/8-01sc-classical-mechanics-fall-2016/pages/review-vectors/}{Classical Mechanics}} have concepts that are presented as pre-segmented short videos.
Here, we re-assemble the small segments to build the original complete lecture.
We trim the intro and outro from short video clips to avoid biasing the models to identify the segments easily.

\section{Lecture Segmentation}

Our lecture segmentation approach involves three stages (Fig.~\ref{fig:banner_2}).
In the first stage, we extract features from diverse modalities of the lecture (Sec.~\ref{subsec:feature_extraction} and Fig.~\ref{fig:banner_2}a).
In the second stage, we learn lecture-aware representations by aligning the visual content with the corresponding narration using self-supervision (Sec.~\ref{subsec:learning_embeddings} and Fig.~\ref{fig:banner_2}b).
Finally, we perform segmentation using TW-FINCH~\cite{sarfraz2021temporally} on the learned representations (Sec.~\ref{subsec:tw_finch} and Fig.~\ref{fig:banner_2}c).

\subsection{Video clip feature extraction}
\label{subsec:feature_extraction}

We divide a lecture into small clips of 10-15 seconds while ensuring that subtitles are not split.
This clip is a basic unit for segmentation, \ie~segmentation boundaries can be placed before or after, not in between.
The chosen duration is small enough to not introduce boundary errors for segmentation but big enough to contain meaningful information about the lecture, as will also be shown empirically.

\paragraph{Video feature extraction.}
\label{subsubsec:video_feature_extraction}
The visual clip representation consists of three feature types: OCR, 2D, and 3D.
The \emph{OCR feature} encodes the output text from an OCR API using the BERT sentence transformer model.
Specifically, we use MPNet \texttt{(all-mpnet-base-v2)}~\cite{song2020mpnet, wolf-etal-2020-transformers} from HuggingFace to obtain a 768-dimensional vector that captures the semantic information of the recognized text.
The \emph{2D and 3D features} are extracted using a video feature extraction pipeline~\cite{miech2019howto100m}.
An ImageNet pre-trained Resnet-152~\cite{he2016deep} model produces 2D features at 1 fps while
the 3D features are extracted using the Kinetics~\cite{carreira2017quo} pre-trained ResNeXt-101~\cite{hara2018can} to obtain 1.5 features per second.
We apply max-pooling across the temporal dimension to obtain 2048-dimensional vectors, $\mathbf{v_{2d}}$ and $\mathbf{v_{3d}}$ respectively.

\paragraph{Text feature extraction}
\label{subsubsec:text_feature_extraction}
uses the same model as used for OCR.
The text feature encodes the instructor's spoken words or subtitles corresponding to each video clip.

\subsection{Learning joint text-video embeddings}
\label{subsec:learning_embeddings}

Our approach transforms features from off-the-shelf models
into lecture-aware embeddings
and is inspired by popular works on instructional videos~\cite{miech2019howto100m, rouditchenko2020avlnet}.

\paragraph{Model architecture.}
Fig.~\ref{fig:banner_2}b depicts our model used to learn lecture-aware embeddings by matching the visual feature of a clip with its corresponding text pair.
We first extract the visual and textual features for a video clip $C$ and transcript (text) $T$ using the feature extraction pipelines described above.
We pass the OCR feature through a fully-connected layer to obtain a 2048-dimensional vector $\mathbf{o}$, and concatenate it with $\mathbf{v_{2d}}$ and $\mathbf{v_{3d}}$ to form a 6144-dimensional vector $\mathbf{c}$ describing the clip $C$.
Similarly, the text feature vector (output of the transformer) is passed through a fully connected layer to obtain a 4096-dimensional vector $\mathbf{t}$, representing text $T$.
Next, we learn a projection using the non-linear context gating~\cite{miech2018learning, miech2019howto100m} defined as follows:
\begin{equation}
f(\mathbf{c}) = (W_1^c \mathbf{c} \: + \: b_1^c) \:\: \odot \:\: \sigma(W_2^c (W_1^c \mathbf{c} \: + \: b_1^c) \: + \: b_2^c) \, ,
\end{equation}
\vspace{-8mm}
\begin{equation}
g(\mathbf{t}) = (W_1^t \mathbf{t} \; + \; b_1^t) \:\: \odot \:\: \sigma(W_2^t (W_1^t \mathbf{t} \; + \; b_1^t) \; + \; b_2^t) \, ,
\end{equation}
where $W_1^c, W_2^c, W_1^t, W_2^t$ and $b_1^c, b_2^c, b_1^t, b_2^t$ are learnable parameters, $\odot$ is element-wise multiplication and $\sigma$ is an element-wise sigmoid.
$f(\mathbf{c})$ and $g(\mathbf{t})$ are 4096-dimensional embeddings, which are used later for the segmentation task.

\paragraph{Loss function.}
We train our embedding model's parameters with the max-margin ranking loss~\cite{karpathy2014deep, wang2016learning}.
Specifically, we consider the (cosine) similarity score between a clip $C_i$ and transcript $T_j$ as $s_{ij} = \langle f(\mathbf{c}_i), g(\mathbf{t}_j) \rangle$.
We loop over paired samples of a mini-batch $\mathcal{B}$ and compute the loss as
\begin{equation}
\sum_{i \in \mathcal{B}} \sum_{j \in \mathcal{N}(i)} \max(0, \delta + s_{ij} - s_{ii}) + \max(0, \delta + s_{ji} - s_{ii}) \, ,
\end{equation}
where $s_{ii}$ corresponds to a positive (aligned) clip-transcript pair $(C_i, T_i)$ and should score high, while $\mathcal{N}(i)$ is the set of negative pairs such that half the negative pairs are from the same lecture and act as hard negatives, while the others stem from other lectures~\cite{anne2017localizing, miech2019howto100m}.
Our mini-batch size is $|\mathcal{B}| = 32$ and the margin is set at $\delta = 0.1$.

\subsection{Lecture segmentation with learned embeddings}
\label{subsec:tw_finch}
We extract clip and transcript embeddings from our joint text-video model and concatenate them to obtain an overall representation $\phi_i = [f(\mathbf{c}_i), g(\mathbf{t}_i)]$.
All such representations of a lecture with $N$ clips, $\{\phi_1, \ldots, \phi_N\}$, are passed to the TW-FINCH algorithm~\cite{sarfraz2021temporally} that encodes feature similarity and temporal proximity as a 1-nearest-neighbor graph and produces a clustering as shown in Fig.~\ref{fig:banner_2}c.
Specifically, we denote the feature similarity between clips as $E_s$ and temporal proximity as $E_\tau$. 
\begin{equation}
\label{E_s}
E_s(m, n)=
\begin{cases}
1 - \langle \phi_m, \phi_n \rangle & \text{if } m \neq n \, , \\
1 & \text{otherwise} \, .
\end{cases}
\end{equation}
\begin{equation}
\label{Et}
E_\tau(m, n)=
\begin{cases}
|\tau_m - \tau_n| / T & \text{if } m \neq n \, , \\
1 & \text{otherwise} \, ,
\end{cases}
\end{equation}
where $m, n \in [1, \ldots, N]$, $\tau_m$ and $\tau_n$ are timestamps for the clips $m$ and $n$ and $T$ is the total lecture duration.

We construct a fully-connected graph $\mathcal{G}$ with $N$ nodes that have edge distances obtained as a combination of feature-space distances and temporal proximity
\begin{equation}
\label{eq:g_e_distances}
E(m, n) = E_s(m, n) \cdot E^\alpha_\tau(m, n) \, ,
\end{equation}
where $\alpha$ acts as a further modulating factor.
The graph $\mathcal{G}$ is converted to a 1-nearest-neighbor graph by keeping only one edge to the \emph{nearest} node for each node based on the edge distances defined in $E$, resulting in the first clustering partition.
TW-FINCH~\cite{sarfraz2021temporally} operates recursively and merges clusters (nodes) by averaging their representations and timestamps until the desired number of clusters (connected components) is obtained.
For more details, we request the reader to refer to Algorithm 1 and 2 in~\cite{sarfraz2021temporally}.

Note that the original algorithm~\cite{sarfraz2021temporally} does not include an $\alpha$ scaling factor, or considers it to be 1 (\cf~Eq.~\ref{eq:g_e_distances}).
However, we observed a few cases where this is unable to produce temporally consistent segments using our learned embeddings.
As higher values of alpha amplify the strength of the temporal proximity factor, incrementing it progressively (\eg~by 0.1 steps) yields temporally consistent clusters.

\section{Experiments}

\begin{table*}[t]
\centering
\begin{tabular}{llcccccccc}
\toprule
& & \multicolumn{3}{c}{Feature modality} & & & & \\
& \textbf{Method} & \textbf{visual} & \textbf{textual} & \textbf{learned} & \textbf{NMI $\uparrow$} & \textbf{MOF $\uparrow$} & \textbf{IOU $\uparrow$} & \textbf{F1 $\uparrow$} & \textbf{BS@30 $\uparrow$} \\

\midrule
1 & Na\"{i}ve (Equal Splits) & - & - & - & 71.8 & 75.5 & 62.7 & 74.0 & 32.5\\
2 & Content-Aware Detector~\cite{pysd} & \cmark & - & - & 72.9 & 73.3 & 59.4 & 65.9 & 57.0\\
3 & Text Tiling~\cite{nltk} & - & \cmark & - & 67.9 & 64.7 & 46.3 & 50.9 & 33.7 \\
4 & LDA~\cite{blei2003latent} & - & \cmark & - & 70.0 & 72.4 & 57.6 & 68.2 & 38.8 \\
5 & K-Means & - & - & \cmark & 63.9 & 66.8 & 48.2 & 55.7 & 44.9 \\
6 & CTE~\cite{kukleva2019unsupervised} & - & - & \cmark & 67.2 & 67.3 & 48.1 & 57.3 & 41.5 \\
\midrule
7 & & \cmark & - & - & 71.6 & 71.3 & 56.5 & 66.4 & 46.9\\
8 & Vanilla TW-FINCH~\cite{sarfraz2021temporally} & - & \cmark & - & 74.6 & 75.4 & 62.0 & 71.2 & 48.9\\
9 & & \cmark & \cmark & - & 74.9 & 75.1 & 61.7 & 70.9 & 52.1\\
\midrule
10 & \textbf{Ours} & - & - & \cmark & \textbf{79.8} & \textbf{80.3} & \textbf{69.2} & \textbf{76.9} & \textbf{58.7}\\
\bottomrule
\end{tabular}
\vspace{-2mm}
\caption{Segmentation performance on all 350 lectures from 15 courses. 
Our approach outperforms all baselines.
Here, \textit{\underline{learned} feature modality} refers to the features extracted from our joint text-video embedding model (Sec.~\ref{subsec:learning_embeddings}).
For rows 2-4, the \textit{\underline{visual} and \underline{textual} feature modalities} refer to the unprocessed lecture video or transcripts respectively.
For rows 7-9, \textit{\underline{visual} and \underline{textual} feature modalities} refer to the features obtained from pre-trained backbones (ResNet or BERT, Sec.~\ref{subsec:feature_extraction}).}
\vspace{-2mm}
\label{table:baselineComp}
\end{table*}

We evaluate our proposed approach for lecture segmentation and present extensive ablation studies.

\subsection{Experiment setup}
\paragraph{Training procedure}
involves two stages.
In the first stage, we pre-train the embedding model (Sec.~\ref{subsec:learning_embeddings}) on the Courses without Segmentation (CwoS).
In the second stage, we fine-tune our embedding model on the Courses with Segmentation (CwS) in an unsupervised manner.
Note that we do not update the feature extraction backbones (BERT, ResNet, \etc).
Next, we extract the visual and textual embeddings from the trained model, which are used to perform segmentation using the TW-FINCH algorithm.
We evaluate the segments obtained from TW-FINCH using five different metrics described below.
Additional training details can be found in the supplementary material (Sec.~\ref{sec:training_details}).

\paragraph{Evaluation dataset.}
We evaluate all 15 courses of CwS to report performance.
Our self-supervised fine-tuning process can be easily extended to a new course that needs segmentation.
Further impact of pre-training and fine-tuning strategies is evaluated in Sec.~\ref{sec:ablation}, Ablation 2.

\paragraph{Evaluation metrics.}
Normalized Mutual Information (NMI) is a standard clustering metric~\cite{manning_irbook};
Mean over Frames (MoF), F1-score, and Intersection over union (IoU) or the Jaccard index are standard metrics used in segmentation (\eg~\cite{sarfraz2021temporally}); and
Boundary Score @ $k$ (BS@$k$), is the average number of predicted boundaries matching with the ground truth boundaries within a $k$ second interval.
Different from the above metrics, BS@$k$ measures the localization of boundaries rather than the overlap of segments.

\subsection{Comparison against Segmentation Baselines}

We briefly describe the baselines below:

\paragraph{1. Na\"{i}ve.}
The video lecture is split into equal parts based on the number of ground-truth (GT) segments.

\paragraph{2. Content-Aware Detector}~\cite{pysd}
is a shot/scene detection algorithm that detects jump cuts in a video by finding areas of high difference between two adjacent frames.
While there is no direct way to set the number of segments,
we search across several thresholds to generate the GT number of segments to ensure a fair comparison.

\paragraph{3. Text Tiling}
utilizes only the transcripts to predict the segments.
We implement text tiling using the NLTK~\cite{nltk} library. 
As there is no way to set the number of clusters, we let the algorithm decide the appropriate number of clusters.

\paragraph{4. Latent Dirichlet Allocation (LDA)}~\cite{mallet,blei2003latent} 
is a generative probabilistic model that automatically discovers hidden topics based on a text corpora.
LDA is used as a baseline in identifying
topic transitions in educational videos~\cite{gandhi2015topic} and many other topic modeling works~\cite{bianchi2021pre, terragni2021octis}.
We train the LDA model on the transcripts of AVLectures and represent each clip
as a distribution over topics.
Finally, we use TW-FINCH to perform lecture segmentation using these vectors.

\paragraph{5. K-Means}
clustering algorithm is applied to the learned embeddings from our joint text-video embedding model.

\paragraph{6. CTE}~\cite{kukleva2019unsupervised}
is a \emph{strong unsupervised approach} that infuses features with relative temporal information and clusters them using K-Means.
We report CTE scores using learned embeddings from our joint model.

\paragraph{7. Vanilla TW-FINCH}~\cite{sarfraz2021temporally}.
Visual and textual features from the feature extraction pipeline in Sec.~\ref{subsec:feature_extraction} are adopted here (no lecture-awareness).
We apply the TW-FINCH segmentation algorithm directly on these features.

\begin{figure}[t]
\includegraphics[width=\linewidth]{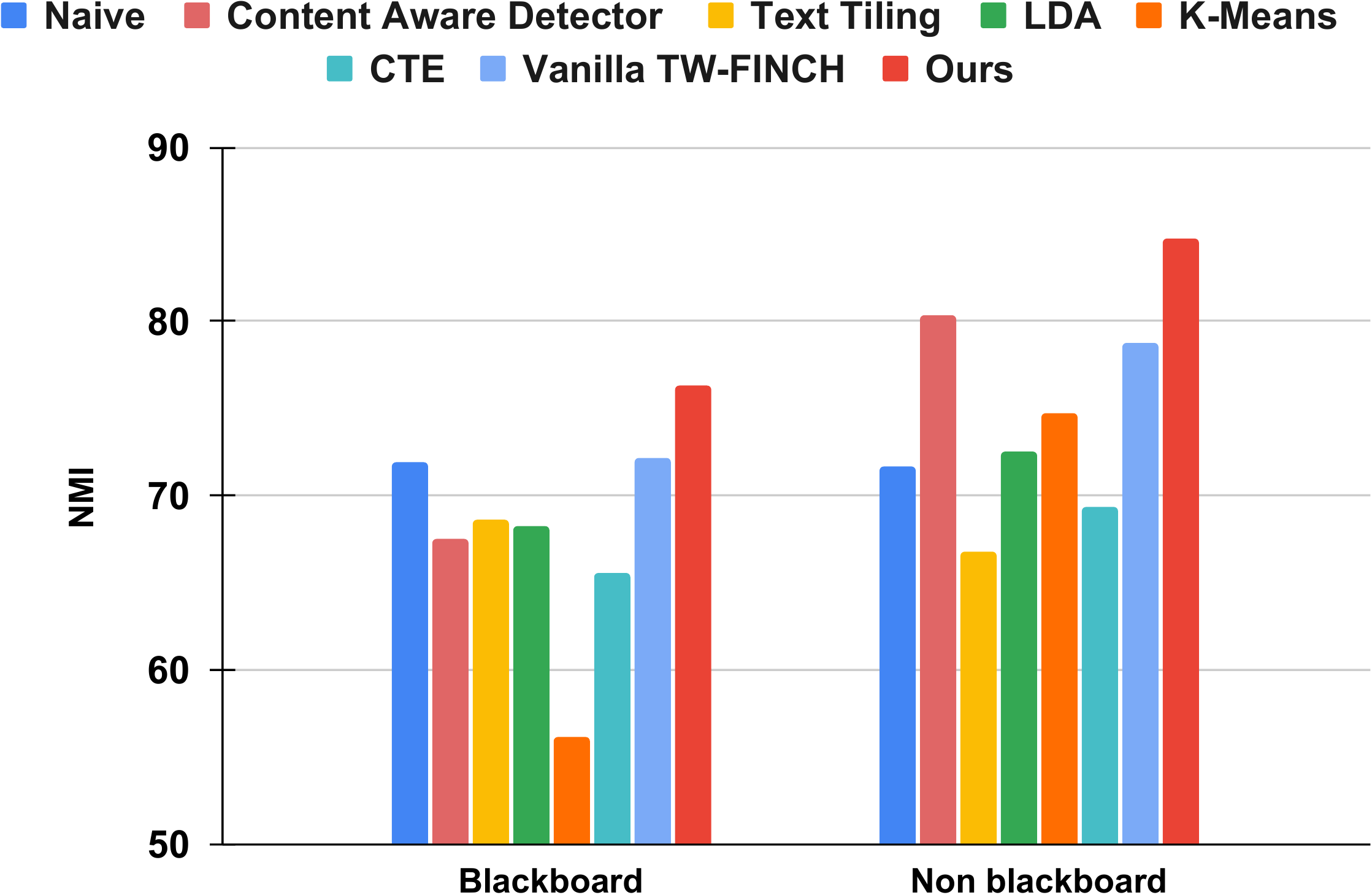}
\vspace{-5mm}
\caption{Comparing NMI across all methods grouped by presentation mode: blackboard and non-blackboard.}
\label{fig:course_wise_score}
\end{figure}

\vspace{1mm}
We compare all baselines against our approach and report performance in Table~\ref{table:baselineComp}.
For K-Means (row 5) and CTE (row 6), we report the best performance with learned features, while detailed ablations are presented in the Sec.~\ref{sec:supp_ablation} of the supp. mat.
We observe that the Na\"{i}ve baseline (row 1) performs quite well, and in fact outperforms strong baselines with learned features such as K-Means (row 5) and CTE (row 6).
This may be due to an inherent bias of the instructor spending close to equal amounts of time on various sub-topics of the lecture (supp. mat. Sec.~\ref{sec:supp_ablation} digs deeper into this).
The text-only approach, Text Tiling (row 3) lags behind the visual-only approach Content-Aware Detector (row 2) as the latter performs specially well on non-blackboard courses (see Fig.~\ref{fig:course_wise_score}).
An additional factor is that we are unable to select the ground-truth number of clusters for Text Tiling.
Our approach (row 10) outperforms all baselines.
In fact, the gap between our approach and Vanilla TW-FINCH baselines (rows 7-9) highlights the importance of training lecture-aware representations using the joint text-video embedding model, as even a combination of both modalities (row 9) falls short of our approach by almost 5\% on NMI.
This emphasizes the importance of learning lecture-aware embeddings in a self-supervised manner.

\begin{figure}[t]
\includegraphics[width=\linewidth]{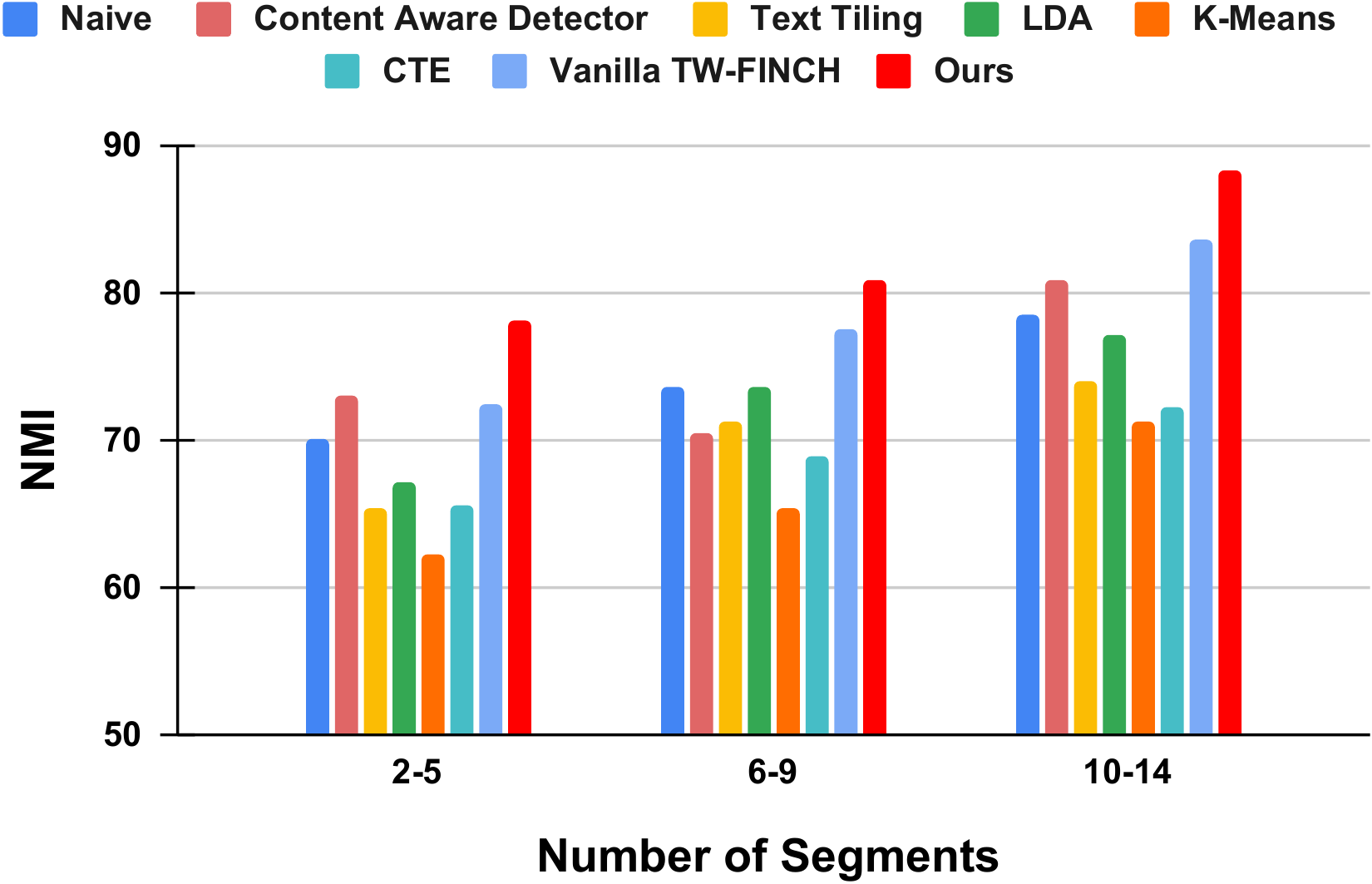}
\vspace{-5mm}
\caption{Comparing NMI across all methods grouped by the number of ground-truth segments.}
\label{fig:segment_wise_score}
\end{figure}

We further analyze the results by slicing lectures based on the number of GT segments in Fig.~\ref{fig:segment_wise_score}.
Our method outperforms all the other baselines irrespective of the number of segments in the ground truth, indicating the robustness of our approach.
Another way is to slice the data based on presentation mode, specifically blackboard and non-blackboard.
Fig.~\ref{fig:course_wise_score} shows a similar trend, our approach outperforms all baselines in both scenarios.
Interestingly, the Na\"{i}ve baseline works well for blackboard lectures (perhaps indicative of relatively equal time allocation across sub-topics), while slide-based lectures with clear transitions are segmented well by the visual Content-Aware Detector.

\subsection{Ablation Studies}
\label{sec:ablation}
We present various ablation studies to understand the contributing factors to our approach's performance.

\paragraph{1. How important is each visual feature?}
To understand the impact of each individual visual feature, we train separate models on all combinations of visual features and report performance in Table~\ref{table:feature_comp}.
We observe that although the individual features perform reasonably well, OCR outperforms 2D and 3D representations, and it is the combination of all features that outperforms all other variations.

\begin{table}[t]
\small
\centering
\tabcolsep=0.13cm
\begin{tabular}{cccccccc}
\toprule
\multicolumn{3}{c}{Features} & \multicolumn{5}{c}{Metrics} \\
\textbf{2D} & \textbf{3D} & \textbf{OCR} & \textbf{NMI $\uparrow$} & \textbf{MOF $\uparrow$} & \textbf{IOU $\uparrow$} & \textbf{F1 $\uparrow$} & \textbf{BS@30 $\uparrow$} \\

\midrule
\cmark & - & - & 76.6 & 76.8 & 64.4 & 73.0 & 54.4\\
- & \cmark & - & 75.1 & 76.0 & 62.9 & 72.2 & 50.7\\
- & - & \cmark & 78.9 & 79.7 & 68.2 & 76.2 & 57.7\\
\cmark & \cmark & - & 76.6 & 77.0 & 64.7 & 73.5 & 53.9\\
\cmark & - & \cmark & 79.5 & 80.3 & 69.1 & 76.9 & 58.6\\
- & \cmark & \cmark & 78.4 & 79.5 & 68.3 & 76.4 & 57.9\\
\cmark & \cmark & \cmark & \textbf{79.8} & \textbf{80.3} & \textbf{69.2} & \textbf{76.9} & \textbf{58.7}\\
\bottomrule
\end{tabular}
\vspace{-2mm}
\caption{Impact of visual features.}
\label{table:feature_comp}
\vspace{-5mm}
\end{table}

\begin{figure*}[t]
\includegraphics[width=\linewidth]{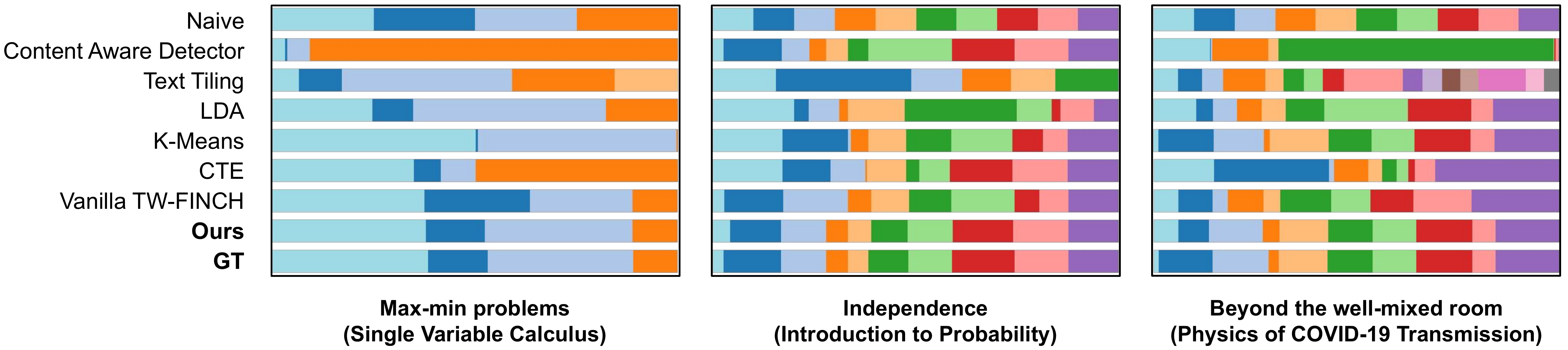}
\vspace{-6mm}
\caption{Segmentation examples for three lectures. Our approach closely resembles the ground-truth. Best viewed in color.
}
\label{fig:qualitative_seg}
\vspace{-4mm}
\end{figure*}

\begin{table}[h]
\small
\tabcolsep=0.10cm
\centering
\begin{tabular}{lccccccc}
\toprule
& \textbf{PT} & \textbf{FT} & \textbf{NMI $\uparrow$} & \textbf{MOF $\uparrow$} & \textbf{IOU $\uparrow$} & \textbf{F1 $\uparrow$} & \textbf{BS@30 $\uparrow$} \\
\midrule

1 & HowTo100M & - & 73.0 & 58.8 & 68.3 & 73.0 & 48.5\\
2 & HowTo100M & CwS & 74.5 & 75.1 & 61.5 & 71.0 & 49.7\\
3 & - & CwS & 78.5 & 79.0 & 67.2 & 75.3 & 57.2\\
4 & CwoS & - & 77.7 & 78.0 & 66.0 & 74.2 & 57.1 \\
5 & \textbf{CwoS} &  \textbf{CwS} & \textbf{79.8} &  \textbf{80.3} &  \textbf{69.2} &  \textbf{76.9} &  \textbf{58.7}\\
\bottomrule
\end{tabular}
\vspace{-2mm}
\caption{Impact of pre-training (PT) on HowTo100M or CwoS. The second column indicates whether unsupervised fine-tuning (FT) is performed on CwS.}
\label{table:pretrain_finetune}
\vspace{-2mm}
\end{table}

\paragraph{2. Impact of training datasets.}
Educational lecture videos are very different compared to instructional videos or movies. Lecture videos typically have much less dynamic visual content and compensate for this through substantial amounts of textual information, both accompanying (narrated speech/transcripts) and even inside the video (which we extract using OCR). As a result, the representations learned from instructional videos may not transfer well to the tasks in the education domain, necessitating a collection of lecture videos for learning representations.

We validate the above claim by showing that pre-training on AVLectures is more effective than pre-training on the general instructional videos (\eg~HowTo100M) for the lecture segmentation task, see Table~\ref{table:pretrain_finetune}.
While using a model to improve representations is clearly better than the na\"{i}ve baseline (NMI 73.0 \vs~71.8), we can see that a model pre-trained on AVLectures (rows 3-5) outperforms a model pre-trained on HowTo100M (rows 1-2) consistently.
This strengthens our dataset contribution and highlights the importance of pre-training on AVLectures for tasks in the education domain.
In row 4, though the model is trained only on CwoS, it is able to generalize well to unseen courses and predict reasonable segmentation boundaries.
After fine-tuning the model on CwS we get a slight boost in performance (row 5).
Row 5 outperforms row 3 that is trained only on CwS, justifying our adoption of pre-training on CwoS followed by fine-tuning on CwS.
Note that all the training is performed in an unsupervised manner and only applies to the text-video embedding model.

\begin{table}[t]
\small
\tabcolsep=0.12cm
\centering
\begin{tabular}{lccccc}
\toprule
\textbf{Embed. type} & \textbf{NMI $\uparrow$} & \textbf{MOF $\uparrow$} & \textbf{IOU $\uparrow$} & \textbf{F1 $\uparrow$} & \textbf{BS@30 $\uparrow$} \\

\midrule

Visual & 78.6 & 79.1 & 67.7 & 75.7 & 57.9\\
Textual & 75.6 & 77.0 & 64.4 & 73.5 & 50.3\\
\textbf{Visual + Textual} &  \textbf{79.8} &  \textbf{80.3} &  \textbf{69.2} &  \textbf{76.9} &  \textbf{58.7}\\
\bottomrule
\end{tabular}
\vspace{-2mm}
\caption{Impact of different embedding modalities.}
\label{table:different_embds}
\vspace{-4mm}
\end{table}

\paragraph{3. Impact of modalities.}
From the joint text-video embedding model we can extract visual and textual embeddings.
We compare visual-only, textual-only, and a concatenation of visual and textual learned embeddings in Table~\ref{table:different_embds}.
A combination of both modalities shows best results.

\begin{table}[t]
\small
\tabcolsep=0.07cm
\centering
\begin{tabular}{lccccccc}
\toprule
\textbf{PT} & \textbf{FT} & \textbf{Duration} & \textbf{NMI $\uparrow$} & \textbf{MOF $\uparrow$} & \textbf{IOU $\uparrow$} & \textbf{F1 $\uparrow$} & \textbf{BS@30 $\uparrow$}\\

\midrule
 & & 4-8 & 53.2 & 58.7 & 53.0 & 40.9 & 26.4 \\
\cmark & - & \textbf{10-15} & \textbf{77.7} & \textbf{78.0} & \textbf{66.0} & \textbf{74.2} & \textbf{57.1} \\
 & & 20-25 & 73.9 & 77.0 & 64.6 & 74.8 & 36.7 \\

\midrule
 & & 4-8 & 54.6 & 60.0 & 54.1 & 42.2 & 26.6 \\

\textbf{\cmark} &  \textbf{\cmark} & \textbf{10-15} & \textbf{79.8} &  \textbf{80.3} &  \textbf{69.2} &  \textbf{76.9} &  \textbf{58.7} \\
 & & 20-25 & 74.5 & 77.7 & 65.6 & 75.6 & 36.8 \\

\bottomrule
\end{tabular}
\vspace{-2mm}
\caption{Performance for different clip durations (in seconds). PT: Pre-training on CwoS, FT: Fine-tuning on CwS.}
\label{table:splits_4s8s_main}
\vspace{-4mm}
\end{table}

\paragraph{4. Impact of lecture clip duration.}
Works on instructional videos such as~\cite{miech2020milnce,miech2019howto100m} typically split videos into short clips of 4s.
We perform an experiment to determine an appropriate clip duration for lecture videos: 4-8s, 10-15s, or 20-25s.
The results reported in Table~\ref{table:splits_4s8s_main} coincide with our expectations that
4-8s clips are too short to capture meaningful information while
20-25s clips are harder to represent due to the pooling operation and also cause a significant drop in BS@30 due to their longer duration.
Clips of 10-15s are a good compromise and span meaningful lecture content while not losing information to pooling.

\paragraph{Additional ablations} on the number of GT segments, max-margin \vs~contrastive loss, different language models, embedding dimension, and evaluation of BS@$k$ at multiple values of $k$ are presented in Sec.~\ref{sec:supp_ablation} of the supp. mat.

\subsection{Qualitative results}
We visualize segmentation outputs for three video lectures from different courses in Fig.~\ref{fig:qualitative_seg} and compare our method with all other baselines.
It is clear that our method yields better segments (overlap) and boundaries as opposed to other methods that produce noisy segments.
In the third lecture, the first and second predicted segments of our approach are different from the GT while the other boundaries are detected correctly.
We explain failure cases in Sec.~\ref{sec:failure_case} and show more results in Sec.~\ref{sec:qualitative_results} of the supp. mat.

\begin{figure}[t]
\centering
  \includegraphics[width=0.85\linewidth]{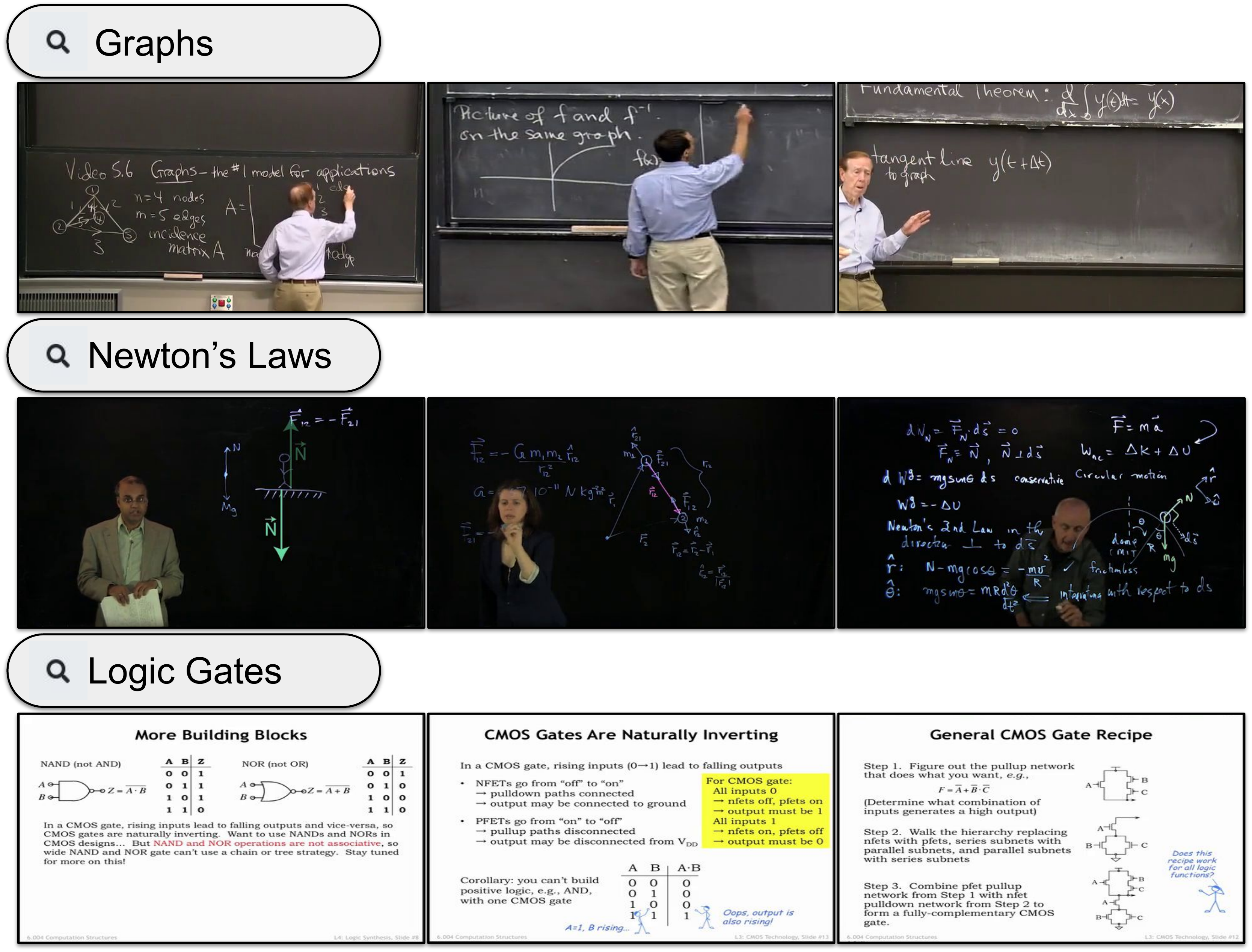}
  \vspace{-2mm}
  \caption{Examples of text-to-video retrieval for different queries using our learned joint embeddings. Our model is able to retrieve relevant lecture clips based on the query.}
  \label{fig:qualitative}
  \vspace{-4mm}
\end{figure}

An additional problem that can be addressed using the embeddings learned from our joint text-video model is the text-to-video retrieval task.
Given a text query, we retrieve a list of lecture clips for which the similarity scores with the text query are the highest.
Fig.~\ref{fig:qualitative} shows some of the retrieved clips for various text queries.
We can see that our model is able to relate the visual notion of graphs with the word.
Similar results are observed for the other queries.
Sec.~\ref{sec:qualitative_results} of the supp. mat. shows many more examples.

\section{Conclusion}

We made two significant contributions.
We introduced \emph{AVLectures}, a large-scale audio-visual lectures dataset sourced from MIT OpenCourseWare, with 86 courses and over 2,350 lectures from various STEM subjects and showed it's efficacy for pre-training on tasks in the educational domain.
We also formulated \emph{unsupervised lecture segmentation} and proposed an approach that learns multimodal representations by matching the narration with temporally aligned visual content.
When used with TW-FINCH, the learned embeddings resulted in significant performance improvements and highlighted the importance of both the visual and the textual modalities.
Thorough experiments demonstrated that our approach outperforms multiple baselines while
comprehensive ablation studies identified the key factors that lead to the success of our approach: textual and visual representations with all 3 features (2D, 3D, OCR) and the pre-training and fine-tuning strategy.

\paragraph{Acknowledgement.}
This material is based upon work supported by the Google Cloud Research Credits program with the award GCP19980904.
We thank MIT-OCW for making their content publicly available.
We thank Vinay Namboodiri for initial discussions on this project.
This work is supported by MeitY, Government of India.

{\small
\bibliographystyle{ieee_fullname}
\bibliography{longstrings,bibiliography}
}

\appendix

\twocolumn[  
    \begin{@twocolumnfalse}
        \begin{center}
             \textbf{
                \Large Unsupervised Audio-Visual Lecture Segmentation \\ \vspace{0.2cm}
                \Large Supplementary Material}
         \end{center}
     \end{@twocolumnfalse}
]

The supplementary material is structured as follows: In Sec.~\ref{sec:wordcloud} we present details about the vocabulary of AVLectures. Next, we analyze a failure case example of segmentation in Sec.~\ref{sec:failure_case}. In Sec.~\ref{sec:manual_segmentation} we provide details on segmenting the lectures manually. Further, we discuss some more ablation studies in Sec.~\ref{sec:supp_ablation} and training details in Sec.~\ref{sec:training_details}. Next, we provide additional qualitative results for both the text-to-video retrieval as well as the lecture segmentation task in section Sec.~\ref{sec:qualitative_results}. Finally, we report segmentation scores for each of the 15 courses in Sec.~\ref{sec:course-wise_scores}.

\section{AVLectures Dataset: Additional Details}
\label{sec:wordcloud}

\begin{figure}[h]
\centering
  \includegraphics[width=\linewidth]{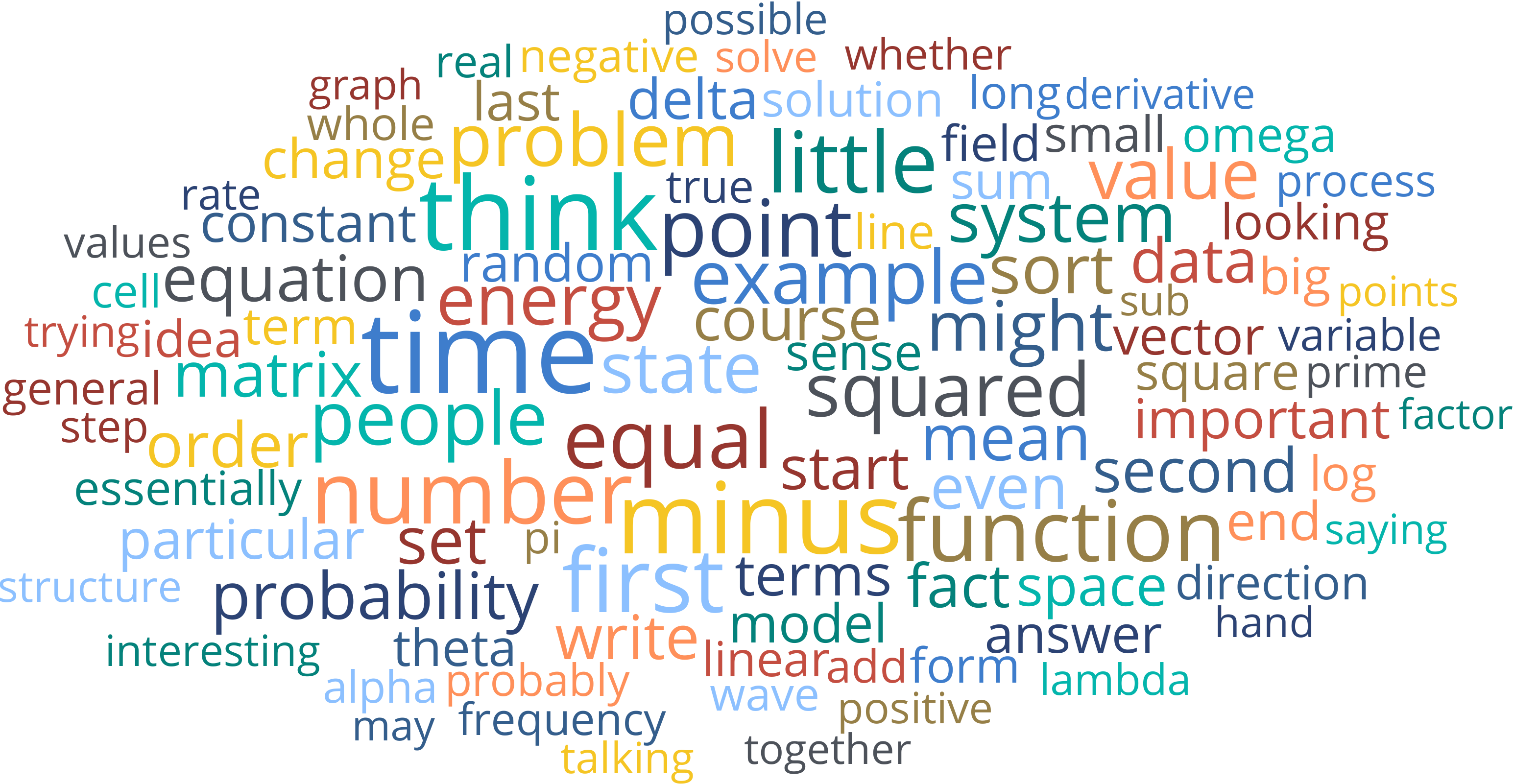}
  \caption{Wordcloud summarizing the distribution of frequently occurring words in AVLectures.}
  \label{fig:wordcloud}
\end{figure}

\begin{figure}[h]
\centering
\includegraphics[width=\linewidth]{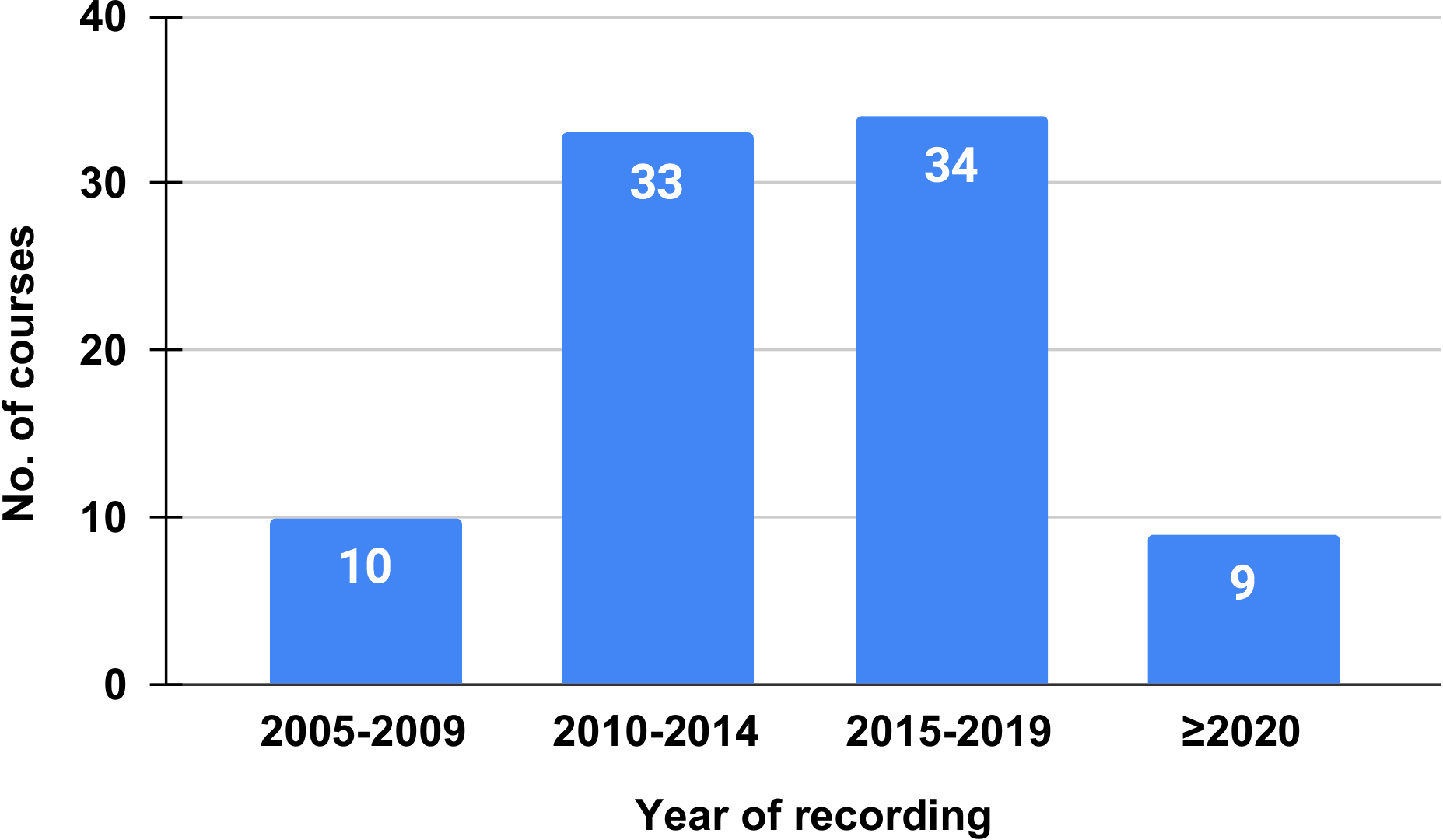}
\caption{Courses from AVLectures that are recorded over the last 2 decades.}
\label{fig:yor}
\end{figure}

AVLectures has a vocabulary size of around 13,000 words with over 7.1M words in total.
Fig.~\ref{fig:wordcloud} shows the distribution of the most occurring words in the dataset. AVLectures is currently dominated by STEM courses, primarily Electrical Engineering \& Computer Science, Physics, and Mathematics, which is evident from the word cloud in Fig.~\ref{fig:wordcloud}.
In our dataset, we have a good mix of old and new courses, as seen in Fig.~\ref{fig:yor}, with the majority being recorded in the last decade.

\section{Deep dive into the failure case}
\label{sec:failure_case}

In this section, we provide more insights into a failure case example of segmentation.
\begin{figure}[h]
\centering
  \includegraphics[width=\linewidth]{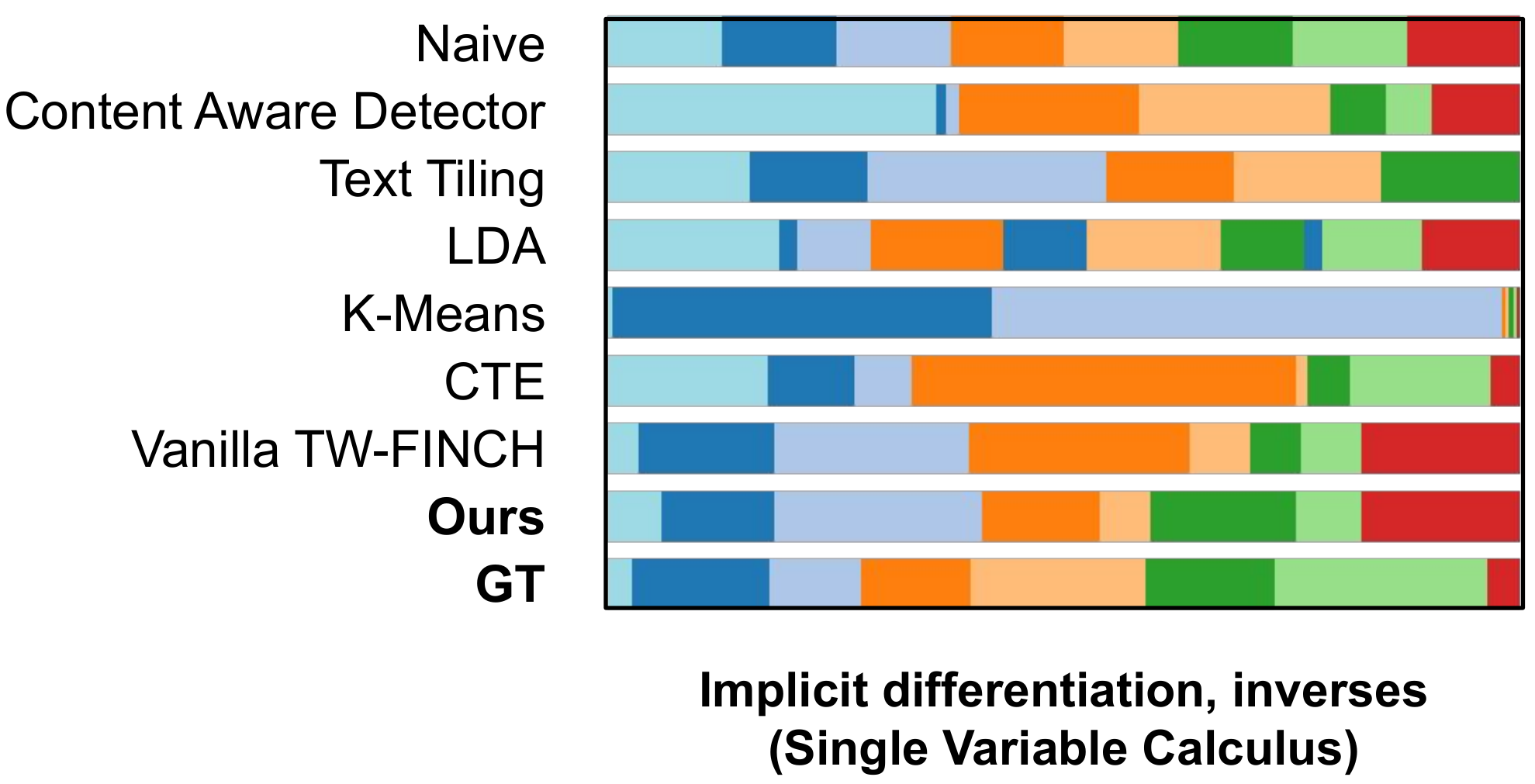}
  \caption{Example of segmentation where the predicted segments differ slightly from the ground truth segments.}
  \label{fig:negative_seg}
\end{figure}

Consider Fig.~\ref{fig:negative_seg}, in which the predicted segments from our model are slightly different from the ground truth segments. However, segmentation is a subjective task, and there can be more than one valid segmentation for some lectures.
Our aim in audio-visual lecture segmentation is to temporally segment a lecture into several smaller segments, such that each segment represents a unique concept/sub-topic. Consider a case in which a single concept can be divided into two smaller concepts. In this case, two valid segmentations are possible: (i) the single concept considered as one complete segment or (ii) the two smaller concepts considered as two separate segments.

Now we compare the Ground Truth (\emph{GT}) segmentation with the segmentation predicted by our model for the lecture \emph{Implicit differentiation} of the \emph{Single Variable Calculus} course\footnote{~\href{https://youtu.be/5q_3FDOkVRQ}{Implicit differentiation lecture video}}. We will refer to the $i^{th}$ segment of Ground Truth as $GT_i$ and that of the segment predicted by our model as $Pred_i$ for the rest of this section. Also, let $len(segment)$ represent the total duration (or the length) of the $segment$. 

$GT_1$ and $Pred_1$ are about the \emph{introduction to implicit differentiation}. However, $Pred_1$ is slightly longer as it includes the part where the \emph{professor greets the late-coming students}. However, this non-lecture segment is a part of $GT_2$ and the rest of it is similar to the $Pred_2$, which covers the topic of \emph{the rational exponent rule}. The ending boundary of $GT_2$ is approximately equal to that of $Pred_2$. Also,

\begin{dmath*}
len(GT_1) + len(GT_2) \approx len(Pred_1) + len(Pred_2)
\end{dmath*}

$GT_3$ discusses the calculation of the \emph{slope of the tangent to a circle using the direct method} and $GT_4$ \emph{using the implicit method}. However, $Pred_3$ combines both the segments into one. The ending boundary of $GT_4$ is close to that of $Pred_3$. Also,

\begin{dmath*}
len(GT_3) + len(GT_4) \approx len(Pred_3)
\end{dmath*}
Next, $GT_5$ is an example involving a \emph{fourth-order equation}. In the case of predicted segmentation, the example is divided into two segments $Pred_4$ and $Pred_5$, which correspond to the two steps involved in solving it. This is an error made by our model as it breaks the two-step solution, however, it is nice to observe that the split is still at a meaningful location. The ending boundary of $GT_5$ is approximately equal to that of $Pred_5$. Also,
\begin{dmath*}
len(GT_5) \approx len(Pred_4) + len(Pred_5)
\end{dmath*}
The last three segments of GT are about the \emph{derivatives of inverse functions and a couple of examples}. 
Among the last three predicted segments, segment 6 and segment 7 are about the \emph{derivatives of inverse functions} and the problem statement of the examples. The final predicted segment covers the solution to both the examples.
\begin{dmath*}
len(GT_6) + len(GT_7) + len(GT_8) \approx len(Pred_6) + len(Pred_7) + len(Pred_8)
\end{dmath*}
Hence, even though the predicted segmentation is slightly different from the GT segmentation, it is still a valid segmentation.

\section{Inter-annotator variation}
\label{sec:manual_segmentation}

\begin{figure}[t]
\centering
  \includegraphics[width=\linewidth]{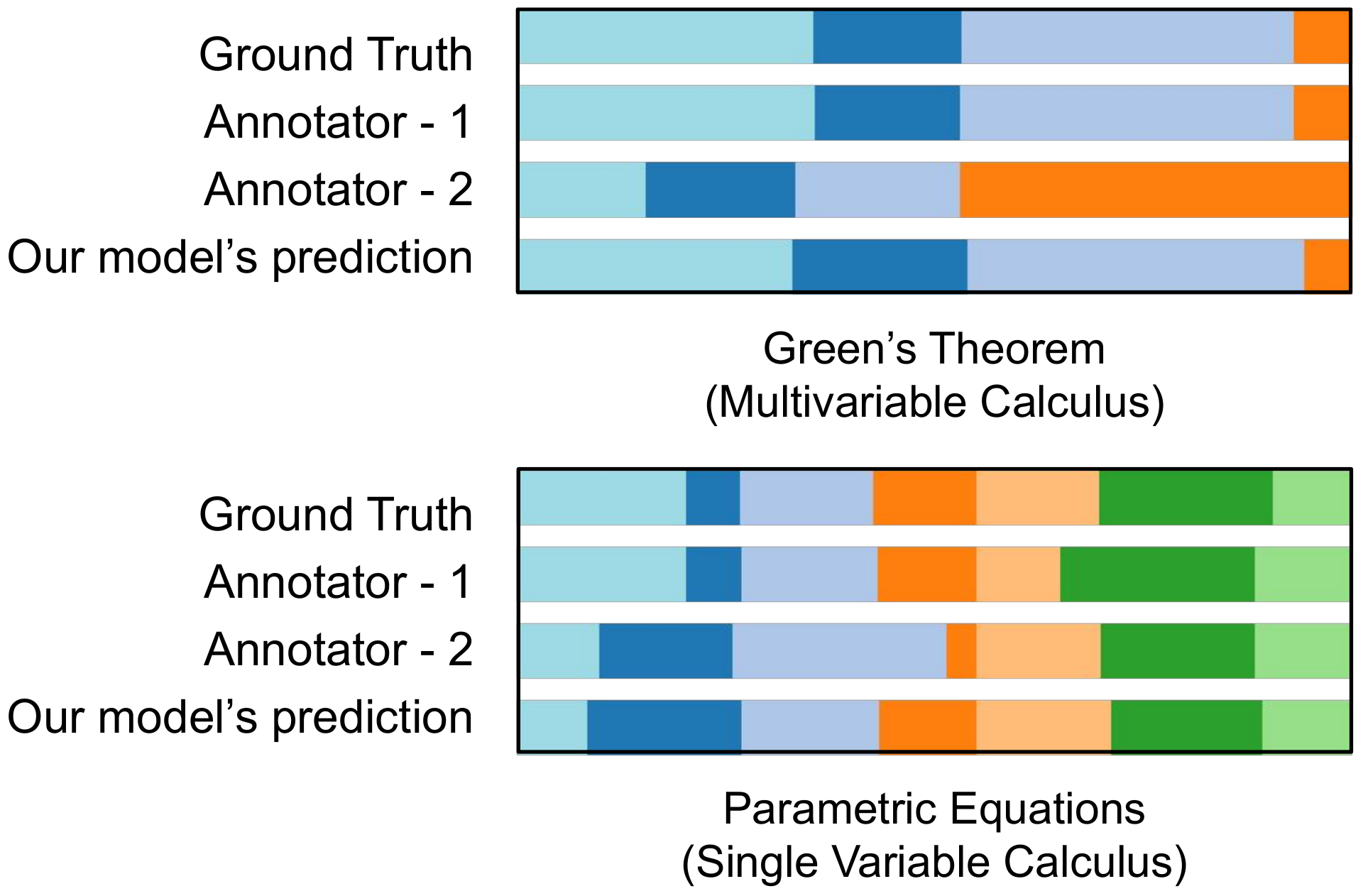}
  \caption{Segmentation of two lectures done manually by two annotators.}
  \label{fig:manual_seg}
\end{figure}

\begin{table}[t]
\centering
\tabcolsep=0.10cm
  \begin{tabular}{lcccccc}
    \toprule
    \textbf{Lecture} & \textbf{Method}& \textbf{NMI} & \textbf{MOF}  & \textbf{IOU} & \textbf{F1} & \textbf{BS@30}\\

    \midrule

    Green's & A-1 & 98.0 & 99.6 & 99.0 & 99.5 & 100.0\\
    Theorem & A-2 & 76.1 & 75.2 & 55.8 & 63.1 & 66.7\\
    & Ours & 86.7 & 95.4 & 88.8 & 94.0 & 66.7\\
    \midrule

    Parametric & A-1 & 89.7 & 92.6 & 87.8 & 93.0 & 66.7\\
    Equations & A-2 & 81.4 & 77.5 & 63.1 & 74.2 & 50.0\\
    & Ours & 86.6 & 84.8 & 76.3 & 84.1 & 66.7\\
    
    \bottomrule

  \end{tabular}
   \caption{Inter-Annotator segmentation scores. Here, A-1 stands for Annotator-1, A-2: Annotator-2, Ours: Our model's prediction.}
  \label{table:manual_seg}
\end{table}

To further analyze the subjective nature of the segmentation task, we asked two annotators to independently segment a few lectures to check the agreement with the corresponding ground truth segmentation and among themselves. Fig.~\ref{fig:manual_seg} shows two such results. In the first example, Annotator-1 considered the topic and its example to be the same segment, whereas Annotator-2 split them into two separate segments. We can see that even though the Annotator-2's segmentation does not match with the Ground Truth, it is still a valid segmentation. Also, our model predicts segments that are closer to that of GT and Annotator-1. In the second example, our model predicts the first two segments in line with Annotator-2's segments while the rest of the segments are similar to that of GT and Annotator-1's segments. 
In Table~\ref{table:manual_seg}, we provide quantitative results of segmentation done by each of the annotators, as well as the prediction from our model with respect to MIT OCW's ground truth.

\section{Additional Ablation studies}
\label{sec:supp_ablation}
\begin{table}[t]
\small
\tabcolsep=0.08cm
\centering
\begin{tabular}{ccccccc}
\toprule
\textbf{Method} & \textbf{Partition} &  \textbf{NMI $\uparrow$} & \textbf{MOF $\uparrow$} & \textbf{IOU $\uparrow$} & \textbf{F1 $\uparrow$} & \textbf{BS@30 $\uparrow$}\\

\midrule

& $2^{\text{nd}}$ last & 63.7 & 61.7 & 59.5 & 42.6 & 42.3 \\
Ours & $3^{\text{rd}}$ last & 72.1 & 59.7 & 39.1 & 42.7 & 65.2\\
& GT & 79.8 & 80.3 & 69.2 & 76.9 & 58.7\\
\midrule

& $2^{\text{nd}}$ last & 58.6 & 58.9 & 54.1 & 40.5 & 27.0 \\
Naive & $3^{\text{rd}}$ last & 66.9 & 51.2 & 33.8 & 39.3 & 38.9\\
& GT & 71.8 & 75.5 & 62.7 & 74.0 & 32.5\\

\bottomrule
\end{tabular}
\caption{Allowing TW-FINCH to estimate the number of clusters.}
\label{table:autocluster_supp}
\end{table}

\paragraph{1. What if the number of segments is unknown?}
It is not trivial to guess the ideal number of segments for the unseen lectures.
In such cases, we let the TW-FINCH algorithm decide the appropriate number of clusters.
TW-FINCH produces a hierarchy of partitions where the number of clusters reduces with successive partitions.
We use the $2^{\text{nd}}$- and the $3^{\text{rd}}$-last partitions to estimate the number of segments automatically and report performance in Table~\ref{table:autocluster_supp}.
We also report scores for the Na\"{i}ve baseline on the above partitions as well.

In addition to the usual metrics we also compute the L1 distance between the ground-truth number of clusters and the number of automatically estimated clusters for both the partitions. The L1 distance between the last and $2^{\text{nd}}$-last partition is 8.554 and that of $3^{\text{rd}}$-last is 4.614.
The $3^{\text{rd}}$-last partition has a lower L1 score compared to the $2^{\text{nd}}$-last partition.
This, along with the other metrics, indicates that the number of clusters generated by the $3^{\text{rd}}$-last partition is closer to the ground-truth.

\begin{table}[h]
\tabcolsep=0.10cm
\centering
  \begin{tabular}{lccccc}
    \toprule
    \textbf{Language Model} & \textbf{NMI $\uparrow$} & \textbf{MOF $\uparrow$} & \textbf{IOU $\uparrow$} & \textbf{F1 $\uparrow$} & \textbf{BS@30 $\uparrow$} \\
    
    \midrule

    Word2Vec & 78.9 & 79.7 & 68.4 & 76.4 & 58.2\\
    mpnet-v1 & 79.1 & 79.7 & 68.3 & 76.2 & 58.4\\
    \textbf{mpnet-v2} &  \textbf{79.8} &  \textbf{80.3} &  \textbf{69.2} &  \textbf{76.9} &  \textbf{58.7}\\

    \bottomrule
  \end{tabular}
\caption{Impact of different Language Models.}
  \label{table:text_embd}
\end{table}

\paragraph{2. Using different language embedding models.}
In this study, we experiment with three different text embeddings, 
\begin{enumerate}
\item{word2vec}: We first preprocess the transcripts by removing the most common stop words. Next, we extract the word embeddings from the GoogleNews pre-trained word2vec model~\cite{mikolov2013efficient}. word2vec encodes each word into to a $300$-dimensional vector.
\item{\texttt{multi-qa-mpnet-base-dot-v1}} (mpnet-v1 in Table~\ref{table:text_embd}): This is a sentence transformer BERT model that uses the pre-trained MPNet~\cite{song2020mpnet} model and is trained on 215M (question, answer) pairs from diverse sources. This model encodes the transcripts into a $768$-dimensional vector. 
\item{\texttt{all-mpnet-base-v2}} (mpnet-v2 in Table~\ref{table:text_embd}): This model uses the pre-trained MPNet~\cite{song2020mpnet} model and is fine-tuned on a 1B sentence pairs dataset using a contrastive learning objective: given a sentence from the sentence pairs, the model should predict which sentence from a randomly sampled other sentences was paired with it.
This is the same model that was described in the Main paper Sec.~\ref{subsec:feature_extraction}.
\end{enumerate}
The results of all three models are reported in Table~\ref{table:text_embd}. Although, the \texttt{all-mpnet-base-v2} model performs slightly better when compared to the other two text embedding models the scores are almost similar in all three variations.
The results show that there is no significant impact on the type of text embeddings that are used to train the model.

\begin{table}[h]
\tabcolsep=0.13cm
\centering
  \begin{tabular}{lccccc}
    \toprule
    \textbf{Embed. dim.} & \textbf{NMI $\uparrow$} & \textbf{MOF $\uparrow$} & \textbf{IOU $\uparrow$} & \textbf{F1 $\uparrow$}  & \textbf{BS@30 $\uparrow$} \\

    \midrule

    512 & 79.3 & 79.7 & 68.3 & 76.1 & \textbf{59.7}\\
    1024 & 79.3 & 80.3 & 68.9 & 76.7 & 59.0\\
    2048 & \textbf{79.8} & \textbf{80.4} & \textbf{69.4} & \textbf{77.1} & 59.6\\
    4096 & \textbf{79.8} & 80.3 & 69.2 & 76.9 & 58.7\\
    \bottomrule

  \end{tabular}
   \caption{Impact of different embedding dimension.}
  \label{table:embedDim}
\end{table}

\paragraph{3. How does the model's embedding dimension affect the performance of segmentation?}
We train the model with four different output embedding dimensions: 512, 1024, 2048 and 4096. It can be seen from Table~\ref{table:embedDim} that the learned features are robust and independent of the feature dimension and therefore has little impact on the overall performance of the model on the segmentation task. Although the embedding dimensions 2048 and 4096 perform slightly better than the rest.

\begin{table}[h]
\small
\tabcolsep=0.06cm
\centering
  \begin{tabular}{lccccccccc}
    \toprule
    & \multicolumn{3}{c}{Feature modality} & & & & & & \\
    & {visual} & {textual} & {learned} & {NMI $\uparrow$} & {MOF $\uparrow$} & {IOU $\uparrow$} & {F1 $\uparrow$} & {BS@30 $\uparrow$} \\
    
    \midrule

    1 & \cmark & - & \xmark & 53.1 & 58.6 & 38.2 & 46.2 & 37.5\\
    2 & - & \cmark & \xmark & 48.5 & 55.1 & 33.5 & 41.0 & 34.3\\
    3 & \cmark & \cmark & \xmark & 53.1 & 58.9 & 38.6 & 46.5 & 37.9\\
    4 & \cmark  & - & \cmark & \textbf{63.9} & \textbf{66.8} & \textbf{48.2} & \textbf{55.7} & \textbf{44.9}\\
    5 & - & \cmark & \cmark & 49.2 & 56.4 & 35.0 & 42.4 & 33.7\\
    6 & \cmark & \cmark & \cmark &  60.2 &  64.9 & 46.0 & 53.3 &  44.1\\

    \bottomrule
  \end{tabular}
\caption{Impact of different feature modalities on K-Means}
  \label{table:kmeans_all}
\end{table}

\begin{table}[h]
\small
\tabcolsep=0.06cm
\centering
  \begin{tabular}{lccccccccc}
    \toprule
    & \multicolumn{3}{c}{Feature modality} & & & & & & \\
    & {visual} & {textual} & {learned} & {NMI $\uparrow$} & {MOF $\uparrow$} & {IOU $\uparrow$} & {F1 $\uparrow$} & {BS@30 $\uparrow$} \\
    
    \midrule

    1 & \cmark & - & \xmark & 65.0 & 65.4 & 45.9 & 55.4 & 38.6\\
    2 & - & \cmark & \xmark & \textbf{67.2} & \textbf{68.1} & \textbf{49.6} & \textbf{59.4} & 35.3\\
    3 & \cmark & \cmark & \xmark & 66.3 & 66.5 & 47.4 & 57.0 & 39.8\\
    4 & \cmark  & - & \cmark & 67.1 & 67.2 & 48.2 & 57.6 & 41.0\\
    5 & - & \cmark & \cmark & 64.7 & 65.7 & 45.4 & 54.8 & 35.6\\
    6 & \cmark & \cmark & \cmark &  \textbf{67.2} &  67.3 & 48.1 & 57.3 &  \textbf{41.5}\\

    \bottomrule
  \end{tabular}
\caption{Impact of different feature modalities on CTE}
  \label{table:cte_all}
\end{table}

\paragraph{4. Impact of different feature modalities on K-Means and CTE~\cite{kukleva2019unsupervised}}
We show the segmentation results for K-means and Continuous Temporal Embedding~\cite{kukleva2019unsupervised} (CTE) on the features extracted using the pipeline (Sec.~\ref{subsec:feature_extraction} Main Paper) as well as on the learned embeddings from our joint text-video model.
The scores are shown in Table~\ref{table:kmeans_all} and~\ref{table:cte_all}.
For K-Means, the learned visual embeddings (row 4) and the combination of learned visual and textual embeddings (row 6) outperforms all other variations by a good margin. 
The results highlight the importance of training lecture-aware representations using our joint text-video embedding model.
For CTE, even though all the scores are relatively closer to each other, the one that uses text features (BERT embeddings) (row 2) and a combination of learned visual and textual embeddings (row 6) perform the best. Note that using a combination of learned visual and textual embeddings results in the highest boundary score, highlighting the importance of our learned representations in predicting better boundaries.

\paragraph{5. Deeper analysis on Na\"{i}ve method performing well.}
As discussed in the paper, one reason why the Na\"{i}ve method is effective is due to an inherent bias of the instructor spending almost equal amounts of time on different topics in certain lectures.
For example, consider a lecture on Multivariate Calculus\footnote{Multivariate Calculus -  ~\href{https://ocw.mit.edu/courses/18-02sc-multivariable-calculus-fall-2010/resources/clip-equations-of-planes-1/}{segment-1}, ~\href{https://ocw.mit.edu/courses/18-02sc-multivariable-calculus-fall-2010/resources/clip-linear-systems-and-planes/}{segment-2}, ~\href{https://ocw.mit.edu/courses/18-02sc-multivariable-calculus-fall-2010/resources/clip-solutions-to-square-systems/}{segment-3}}. Here each of the segment is approximately 16 minutes, thus giving an upper-hand to the naive method. Upon further analysis, we observe that 73 of 350 lectures (nearly 20 \% of CwS) have GT segment boundaries within 3 minutes to the boundaries suggested by the Na\"{i}ve baseline.
We perform an ablation study by varying the number of splits obtained by automatically clustering lectures with TW-FINCH.
The results indicate that splitting lectures at the ground truth number of segments gives a better segmentation performance than splitting it in any other way, as seen in Table~\ref{table:autocluster_supp}.

\begin{figure}[t]
\centering
\includegraphics[width=\linewidth]{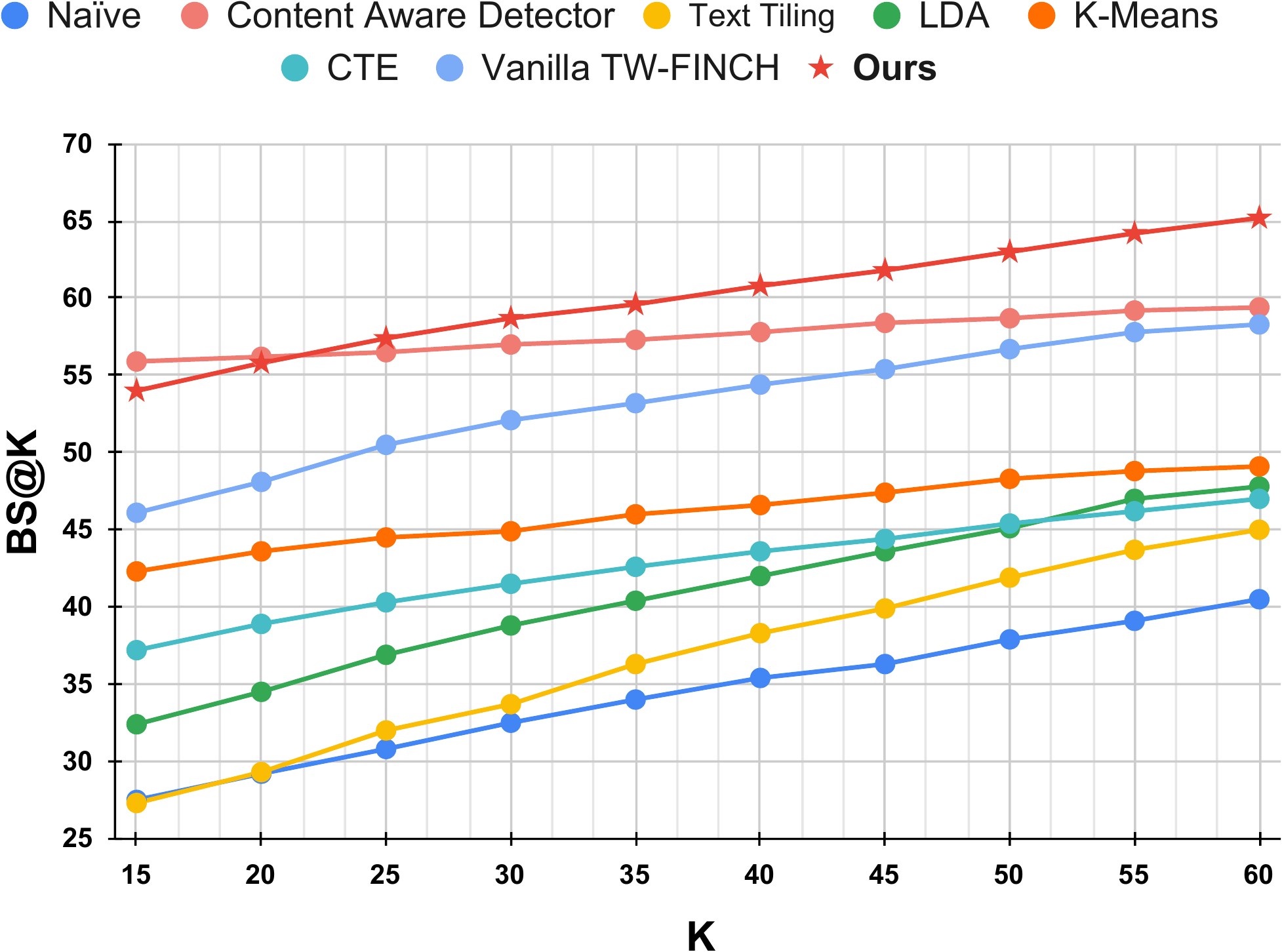}
\caption{Boundary scores at different values of K.}
\label{fig:bs_k_supp}
\end{figure}

\paragraph{6. Boundary scores at various intervals.}
We also perform an ablation study by computing Boundary Scores at various values of K, and it's plot is shown in Fig.~\ref{fig:bs_k_supp}. 
Typically, the instructor spends at least 25-30 seconds (in answering student's questions, erasing the blackboard etc.) before switching to new a topic. This was the reason behind reporting the scores for BS@30 in the paper. As expected, all methods perform worse for lower values of K and as K approaches 15, the use of 10-15s clip sizes hurts performance.

\begin{table}[t]
\small
\tabcolsep=0.2cm
\centering
\begin{tabular}{lccccc}
\toprule
\textbf{Method} & \textbf{NMI $\uparrow$} & \textbf{MOF $\uparrow$} & \textbf{IOU $\uparrow$} & \textbf{F1 $\uparrow$} & \textbf{BS@30 $\uparrow$}\\

\midrule
NCE & 70.6 & 71.5 & 56.3 & 66.3 & 43.2 \\
Ours & \textbf{79.8} & \textbf{80.3} & \textbf{69.2} & \textbf{76.9} & \textbf{58.7}\\

\bottomrule
\end{tabular}
\caption{Segmentation performance when lecture-transcript alignment is done using Noise Contrastive Estimation (NCE) loss.}
\label{table:nce_loss_supp}
\end{table}

\paragraph{7. Impact of lecture-transcript alignment strategies.}
We also compare our approach with a more popular approach that uses Noise Contrastive Estimation (NCE) loss for aligning video-text pairs~\cite{miech2020milnce}. The results are reported in Table~\ref{table:nce_loss_supp}. Our approach, which uses max-margin ranking loss outperforms the NCE loss perhaps due to the scale of the dataset and the limited number of negative samples in the batch. We were unable to train with larger batch sizes due to GPU memory restrictions.

\section{Training details}
\label{sec:training_details}
We train our joint text-video embedding model’s parameters with the max-margin ranking loss. We use a mini-batch size of 32. Our model is trained on a 1080ti NVIDIA GPU using Adam optimizer with a learning rate of 1$e$-4 and a learning rate decay of 0.9. We use the same hyperparameters for both the pre-training and fine-tuning.

\section{Additional Qualitative Results: Retrieval and Segmentation}
\label{sec:qualitative_results}
\begin{figure*}[t]
\centering
  \includegraphics[width=\textwidth]{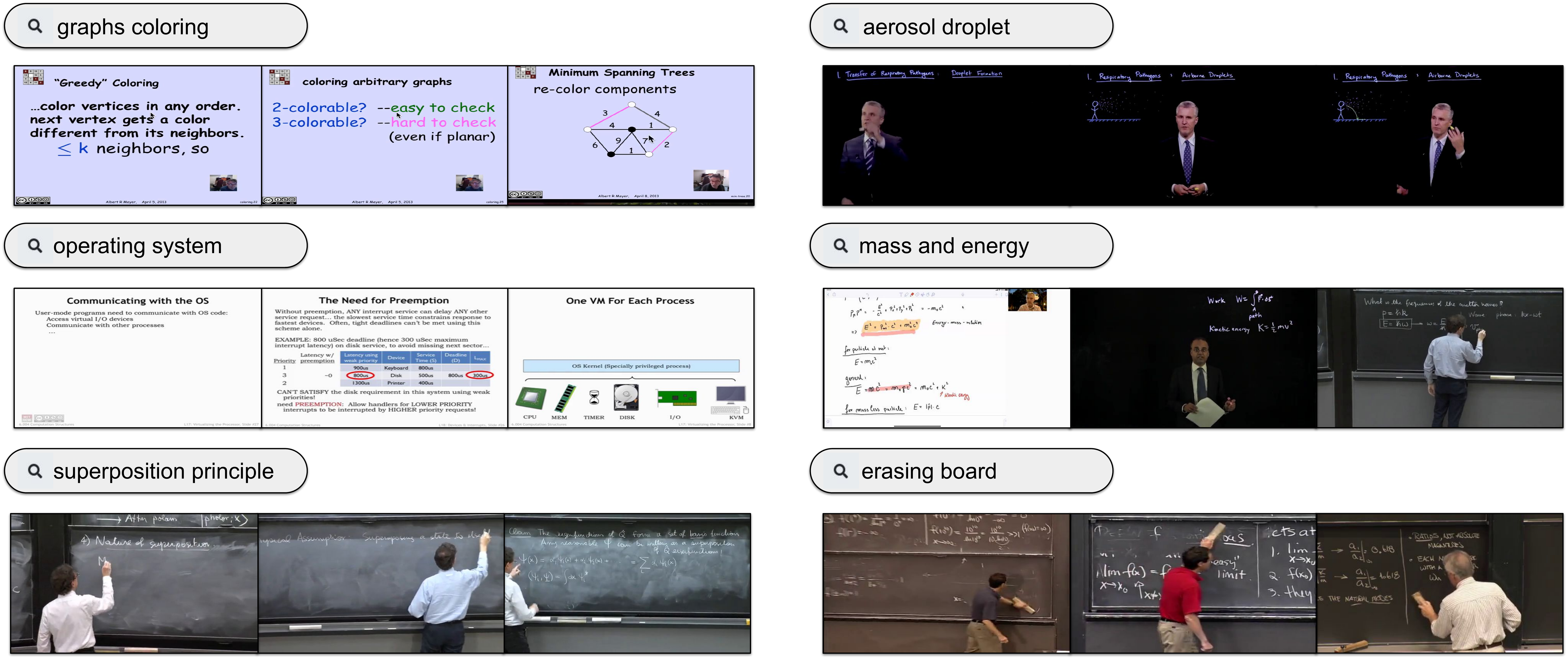}
  \caption{Text-to-video retrieval results on six queries. The figure shows the thumbnails of the top 3 retrieved lecture clips from our model. Our model is able to retrieve relevant lecture clips according to the query.}
  \label{fig:text_to_video}
\end{figure*}

\begin{figure*}[t]
\centering
  \includegraphics[width=\textwidth]{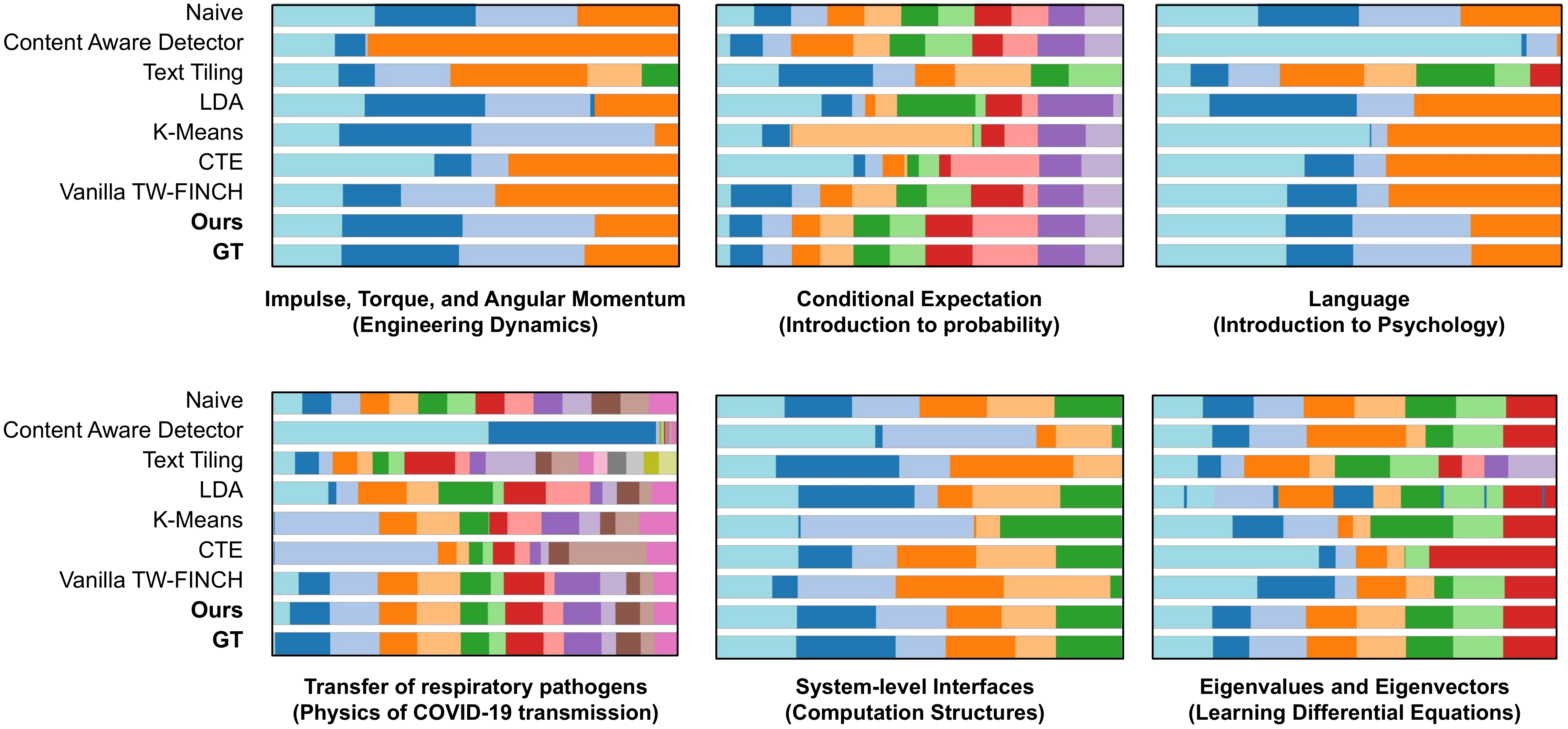}
  \caption{Segmentation examples for six lectures from different courses with varying a number of segments.}
  \label{fig:qual_segmentation}

\end{figure*}

This section shows additional qualitative results for the text-to-video retrieval and the lecture segmentation task. 
Fig.~\ref{fig:text_to_video} shows some of the retrieved clips for different text queries like \emph{graphs coloring}, \emph{operating systems}, etc.
We also tested a query \emph{erasing board} to check the model's comprehension of non-conceptual keywords, as shown in the last example of the figure. Although this query is not present in the transcript, it still correctly retrieves the clips in which the professor erases the blackboard. This demonstrates the importance of pre-training on the CwoS dataset.

Fig.~\ref{fig:qual_segmentation} shows more qualitative results from the lecture segmentation task for lectures from different courses. Regardless of the number of segments, our method yields better segmentation length and boundaries when compared with the other baselines.

\section{Course-wise segmentation results}
\label{sec:course-wise_scores}
\begin{table*}[t]
\centering
\tabcolsep=0.1cm
  \begin{tabular}{ccccccc|ccccccc}
    \toprule
    \textbf{Course ID} & \textbf{Method} & \textbf{NMI} & \textbf{MOF} & \textbf{IOU} & \textbf{F1} & \textbf{BS@30} & \textbf{Course ID} & \textbf{Method} & \textbf{NMI} & \textbf{MOF} & \textbf{IOU} & \textbf{F1} & \textbf{BS@30}\\

    \midrule
    & Na\"{i}ve & 72.1 & 63.9 & 49.0 & 61.6 & 22.2 & 
    & Na\"{i}ve & 66.3 & 78.5 & 66.5 & 77.1 & 39.4\\
    & CAD & 63.2 & 55.9 & 33.4 & 42.7 & 29.0 & 
    & CAD & 51.3 & 66.7 & 47.7 & 58.6 & 35.3\\
    mit001 & LDA & 70.0 & 60.7 & 43.7 & 55.2 & 28.1 & 
    mit002 & LDA & 57.2 & 73.4 & 58.0 & 69.5 & 37.3\\
    & V-TWF & \textbf{76.5} & 68.2 & \textbf{52.5} & \textbf{62.2} & \textbf{45.5} & 
    & V-TWF & 67.8 & 77.4 & 64.7 & 73.7 & 54.1\\
    & Ours & \textbf{76.5} & \textbf{68.4} & 52.3 & \textbf{62.2} & 44.2 & 
    & Ours & \textbf{75.0} & \textbf{83.9} & \textbf{73.9} & \textbf{80.6} & \textbf{57.3}\\
    \midrule

    & Na\"{i}ve & 72.0 & 79.0 & 67.7 & 78.0 & 47.3 & 
    & Na\"{i}ve & 75.8 & 83.2 & 72.1 & 82.8 & 29.8\\
    & CAD & \textbf{96.1} & \textbf{96.0} & \textbf{94.9} & \textbf{94.8} & \textbf{94.8} & 
    & CAD & \textbf{94.9} & \textbf{94.0} & \textbf{89.8} & \textbf{92.1} & \textbf{91.4}\\
    mit032 & LDA & 68.4 & 75.9 & 62.0 & 72.3 & 50.7 & 
    mit035 & LDA & 70.2 & 74.0 & 60.0 & 72.0 & 28.0\\
    & V-TWF & 78.7 & 80.4 & 71.0 & 78.2 & 72.4 &
    & V-TWF & 71.3 & 70.5 & 56.6 & 67.5 & 39.2\\
    & Ours & 87.8 & 88.2 & 81.9 & 86.3 & 86.5 & 
    & Ours & 77.3 & 79.6 & 69.0 & 78.2 & 45.7\\
    \midrule
    
    & Na\"{i}ve & 73.0 & 82.1 & 70.9 & 81.7 & 26.3 & 
    & Na\"{i}ve & 70.1 & 78.9 & 67.1 & 77.4 & 44.8\\
    & CAD & \textbf{98.0} & \textbf{97.1} & \textbf{95.8} & \textbf{96.8} & \textbf{96.7} & 
    & CAD & 76.6 & 77.5 & 61.8 & 66.9 & 70.7\\
    mit038 & LDA & 69.7 & 73.9 & 59.7 & 71.0 & 27.2 & 
    mit039 & LDA & 78.2 & 82.3 & 69.7 & 77.7 & 62.2\\
    & V-TWF & 74.6 & 77.6 & 64.2 & 73.6 & 42.7 &
    & V-TWF & 76.8 & 81.2 & 69.2 & 77.5 & 57.6\\
    & Ours & 76.7 & 78.9 & 66.9 & 75.8 & 45.8 & 
    & Ours & \textbf{83.4} & \textbf{86.0} & \textbf{77.6} & \textbf{83.2} & \textbf{75.6}\\
    \midrule
    
    & Na\"{i}ve & 73.0 & 72.9 & 58.7 & 71.4 & 26.2 & 
    & Na\"{i}ve & 74.4 & \textbf{76.0} & \textbf{63.1} & \textbf{74.5} & 30.0\\
    & CAD & \textbf{94.3} & \textbf{89.5} & \textbf{84.3} & \textbf{86.5} & \textbf{90.1} & 
    & CAD & 57.7 & 56.2 & 35.7 & 44.8 & 26.0\\
    mit049 & LDA & 78.8 & 79.7 & 66.4 & 76.6 & 47.4 & 
    mit057 & LDA & 68.6 & 67.1 & 51.3 & 62.8 & 30.6\\
    & V-TWF & 82.2 & 79.8 & 66.6 & 74.3 & 62.9 & 
    & V-TWF & 71.2 & 69.3 & 53.7 & 65.2 & 32.8\\
    & Ours & 84.4 & 84.7 & 73.8 & 81.4 & 63.2 & 
    & Ours & \textbf{76.3} & \textbf{76.0} & 62.8 & 72.4 & \textbf{41.2}\\
    \midrule
    
    & Na\"{i}ve & \textbf{74.5} & \textbf{77.2} & \textbf{65.0} & \textbf{76.2} & 34.6 &
    & Na\"{i}ve & 73.9 & 72.4 & 58.7 & \textbf{71.4} & 24.1\\
    & CAD & 57.8 & 57.1 & 35.6 & 45.6 & 24.2 & 
    & CAD & 68.3 & 57.5 & 37.4 & 47.4 & 42.1\\
    mit075 & LDA & 74.0 & 73.8 & 59.8 & 70.2 & \textbf{40.3} & 
    mit088 & LDA & 76.9 & 72.2 & 58.2 & 68.6 & 46.0\\
    & V-TWF & 72.2 & 71.5 & 56.7 & 67.6 & 35.4 &
    & V-TWF & 79.2 & 71.7 & 57.2 & 66.1 & 54.2\\
    & Ours & 73.4 & 74.8 & 60.6 & 71.3 & 35.2 & 
    & Ours & \textbf{80.3} & \textbf{74.8} & \textbf{61.8} & 71.0 & \textbf{56.0}\\
    \midrule
    
    & Na\"{i}ve & 65.7 & 81.6 & 70.2 & 80.4 & 43.8 & 
    & Na\"{i}ve & 67.0 & 66.6 & 51.1 & 63.5 & 21.7\\
    & CAD & 63.0 & 75.3 & 61.2 & 68.9 & 58.7 &
    & CAD & 52.0 & 57.9 & 33.6 & 42.1 & 27.5\\
    mit097 & LDA & 65.8 & 79.6 & 66.2 & 74.8 & 56.5 & 
    mit126 & LDA & 67.7 & 68.0 & 52.5 & 63.8 & 24.4\\
    & V-TWF & 72.3 & 81.9 & 71.5 & 79.7 & 67.3 &
    & V-TWF & 69.0 & 69.1 & 51.5 & 62.1 & 39.4\\
    & Ours & \textbf{79.2} & \textbf{86.1} & \textbf{77.5} & \textbf{83.9} & \textbf{72.4} & 
    & Ours & \textbf{72.4} & \textbf{71.1} & \textbf{56.4} & \textbf{66.4} & \textbf{41.5}\\
    \midrule

    & Na\"{i}ve & 76.1 & 65.2 & 47.8 & 60.3 & 23.3 & 
    & Na\"{i}ve & 77.1 & 68.6 & 53.5 & 65.5 & 25.8\\
    & CAD & 76.7 & 66.9 & 59.0 & 60.3 & 63.6 & 
    & CAD & 86.5 & 76.7 & 65.1 & 71.2 & 70.1\\
    mit151 & LDA & 77.0 & 66.9 & 50.4 & 61.0 & 27.4 & 
    mit153 & LDA & 77.7 & 68.6 & 50.7 & 61.1 & 39.8\\
    & V-TWF & 85.2 & 79.4 & 64.0 & 73.2 & 50.4 &
    & V-TWF & 85.3 & 77.5 & 64.2 & 72.0 & 65.4\\
    & Ours & \textbf{95.1} & \textbf{93.6} & \textbf{84.7} & \textbf{88.0} & \textbf{84.6} & 
    & Ours & \textbf{90.8} & \textbf{84.7} & \textbf{74.6} & \textbf{79.9} & \textbf{78.8}\\
    \midrule
    
    & Na\"{i}ve & 81.2 & 85.6 & 76.4 & 85.5 & 31.6 &
    & Na\"{i}ve & 71.8 & 75.5 & 62.7 & 74.0 & 32.5\\
    & CAD & 95.8 & 91.7 & 88.7 & 91.0 & 92.5 &
    Average & CAD & 72.9 & 73.3 & 59.4 & 65.9 & 57.0\\
    mit159 & LDA & 78.6 & 76.8 & 65.9 & 75.6 & 31.3 &
    (across all & LDA & 70.0 & 72.4 & 57.6 & 68.2 & 38.8\\
    & V-TWF & 81.8 & 80.9 & 69.7 & 77.6 & 61.2 &
    the 350 lectures) & V-TWF & 74.9 & 75.1 & 61.7 & 70.9 & 52.1\\
    & Ours & \textbf{98.4} & \textbf{99.4} & \textbf{98.8} & \textbf{99.4} & \textbf{97.2} &
    & \textbf{Ours} & \textbf{79.8} & \textbf{80.3} & \textbf{69.2} & \textbf{76.9} & \textbf{58.7}\\ 
    
    \bottomrule

  \end{tabular}
   \caption{Course-wise segmentation scores. Here, CAD stands for Content Aware Detector, V-TWF : Vanilla TW-FINCH applied on the concatenation of visual and textual features. The last panel shows the average scores across all the 350 lectures of the CwS dataset.}
  \label{table:course_wise}
\end{table*}

\begin{table*}[t]
\centering
\tabcolsep=0.20cm
  \begin{tabular}{lccccc}
    \toprule
    \textbf{Course ID} & \textbf{Course Name}& \textbf{Subject area} & \textbf{\# Lectures}  & \textbf{Avg. \# segments} & \textbf{Mode}\\

    \midrule

    mit001 & Single Variable Calculus & Mathematics & 35 & 7.9 & Blackboard\\
    mit002 & Multivariable Calculus & Mathematics & 35 & 3.2 & Blackboard\\
    mit032 & Classical Mechanics & Physics & 38 & 4.3 & Digital Board\\
    mit035 & Quantum Physics I & Physics & 24 & 4.8 & Blackboard\\
    mit038 & Quantum Physics III & Physics & 24 & 4.2 & Blackboard\\
    mit039 & Introduction to Special Relativity & Physics & 12 & 4.2 & Digital Board\\
    mit049 & Introduction to Nuclear and Particle Physics & Physics & 11 & 6.1 & Digital Board\\
    mit057 & Introduction to Psychology & BCS & 24 & 5.5 & Blackboard\\
    mit075 & Principles of Microeconomics & Economics & 26 & 5.1 & Blackboard\\
    mit088 & Computation Structures & EECS & 21 & 6.6 & Slides\\
    mit097 & Mathematics for Computer Science & EECS & 35 & 3.2 & Slides\\
    mit126 & Engineering Dynamics & ME & 27 & 5.3 & Blackboard\\
    mit151 & Physics of COVID-19 Transmission & Biology & 4 & 9.4 & Digital Board\\
    mit153 & Introduction to Probability & EECS & 26 & 9.3 & Slides\\
    mit159 & Learn Differential Equations & Mathematics & 8 & 6.9 & Blackboard\\
    
    \bottomrule

  \end{tabular}
  \caption{Mapping between course IDs and course names along with additional metadata. Here, BCS stands for Brain and Cognitive Sciences, EECS - Electrical Engineering and Computer Science, and ME - Mechanical Engineering.}
  \label{table:course_stats}
\end{table*}

We report the top 5 segmentation scores for each of the courses of the CwS dataset across all of its lectures in Table~\ref{table:course_wise}.
The mapping between the course ID and the course name is shown in Table~\ref{table:course_stats}, along with other metadata like the subject area, number of lectures, the average number of segments, and the presentation mode. As seen in the table, our method outperforms all of the other baselines for the majority of the courses. However, there are a few courses (mit032, mit035, mit038, and mit049) for which the Content Aware Detector baseline has scores better than the other methods. These are the courses where we combine the individual shorter video segments to form the complete lecture. Since each of these shorter video segments was filmed independently, the lighting/camera angle may have been slightly different for each of these segments. This makes it easier for the Content Aware Detector to predict accurate boundaries. For the other courses, the Content Aware Detector scores are considerably lower than most of the other baselines and our model. All in all, our model outperforms all of the other baselines on an average across all the lectures of the CwS dataset easily, as shown in the last panel of Table~\ref{table:course_wise}.


\end{document}